\documentclass[11pt]{article}

\usepackage[round]{natbib}

\usepackage{amsmath}
\usepackage{amssymb}
\usepackage{amsfonts}
\usepackage{mathrsfs}
\usepackage{mathtools}
\usepackage{hyperref}
\usepackage{fullpage}
\usepackage{amsthm}
\theoremstyle{plain}

\newtheorem{theorem}{Theorem}
\newtheorem{lemma}{Lemma}
\newtheorem{proposition}{Proposition}
\newtheorem{corollary}{Corollary}
\newtheorem{remark}{Remark}

\theoremstyle{definition}
\newtheorem{assumption}{Assumption}
\newtheorem{definition}{Definition}
\newtheorem{example}{Example}

\renewcommand{\d}{\,\mathrm{d}}
\let\div\relax
\DeclareMathOperator{\div}{div}

\newcommand{\A}{\mathcal{A}}
\newcommand{\D}{\mathcal{D}}
\DeclareMathOperator{\EE}{\mathcal{E}}
\let\L\relax
\newcommand{\I}{\mathcal{I}}
\newcommand{\L}{\mathcal{L}}

\DeclarePairedDelimiter{\ang}{\langle}{\rangle}
\DeclarePairedDelimiter{\norm}{\lVert}{\rVert}
\newcommand{\F}{\mathcal{F}}

\newcommand{\Meas}{\mathbb{M}}
\newcommand{\R}{\mathbb{R}}
\let\P\relax
\newcommand{\P}{\mathcal{P}}

\newcommand{\X}{\mathcal{X}}
\DeclareMathOperator{\E}{E}
\DeclareMathOperator{\Var}{Var}
\DeclareMathOperator{\AVar}{AVar}
\newcommand{\smallop}[1][n^{-1/2}]{o_{\mathrm{p}}\left(#1\right)}

\newcommand{\warrow}{\rightsquigarrow}

\title{A unified approach for estimation of unnormalized models\\via Stein operators}
\date{}
\author{Mitsuki Nagai\thanks{Graduate Institute for Advanced Studies, SOKENDAI, e-mail: \texttt{nagai.mitsuki@ism.ac.jp}} \and Keisuke Yano\thanks{The Institute of Statistical Mathematics, e-mail: \texttt{yano@ism.ac.jp}}}

\begin{document}
\maketitle

\begin{abstract}
In this paper, we develop a unified framework connecting proper scoring rules and Stein's method for the estimation of unnormalized models.
We construct a class of proper scoring rules that encompasses score matching, ratio matching, pseudolikelihood, and noise contrastive estimation.
We show that these proper scoring rules naturally induce Stein operators.
This connection yields a new construction principle for Stein operators, 
and enables us to study estimators based on these scoring rules through the geometry associated with the induced Stein operators. 
Using this geometric perspective, we provide a unified
characterization of the asymptotic efficiency of the resulting estimators.
\end{abstract}

\section{Introduction}

Consider estimation in a parametric model
\[
\P \coloneqq \left\{ p_{\theta}(x)=\frac{\tilde{p}_{\theta}(x)}{Z(\theta)} \,\middle|\, \theta\in\Theta \right\}
\]
on a sample space $\X$, where $\Theta\subset\R^d$ and $Z(\theta)\coloneqq\int_{\X}\tilde{p}_{\theta}(x)\d x$ is a normalizing constant.
A normalizing constant is often intractable,
making maximum likelihood estimation difficult in many cases; examples include Markov random fields \citep{Besag1974}, statistical models on manifolds \citep{Mardia1999,Chikuse2003}, Boltzmann machines \citep{Ackley1985}, energy-based models \citep{LeCun2006,Song2021}, and score-based generative models \citep{Lai2025}.

A series of works by \citet{Parry2012}, \citet{Dawid2012}, and \citet{Ehm2012} developed the theory of proper local scoring rules, providing a framework for estimation in unnormalized models that encompasses pseudolikelihood \citep{Besag1975}, score matching \citep{Hyvarinen2005}, and ratio matching \citep{Hyvarinen2007}.

More recently, 
Stein's method has emerged as an 
alternative approach to estimation in unnormalized models \citep{Barp2019,Ebner2025}.
This approach employs a Stein operator $\A_{\theta}$ and a Stein class $\F_{\theta}$ satisfying the characterization
\[
    \E_{\theta^{\star}}[\A_{\theta}f(X)]=0,\quad \forall f\in\F_{\theta}\iff \theta=\theta^{\star}.
\]
The pair $(\A_{\theta},\F_{\theta})$ is called a Stein pair.
Various constructions of Stein pairs have been proposed \citep[e.g.,][]{Gorham2015,Mijoule2023}.

In this paper, we establish a unified approach for a class of proper scoring rules via Stein operators.
Given a suitable linear operator $\L$ together with a concave function $h$, we construct a proper scoring rule that does not depend on the normalizing constant.
Our class of proper scoring rules encompasses score matching, ratio matching, Besag's pseudolikelihood, and noise contrastive estimation \citep{Gutmann2010}.
Moreover, it yields a new construction principle for Stein operators and shows that estimators based on these proper scoring rules can be analyzed through the \textit{geometry} induced by the corresponding Stein operators.
Leveraging this geometric perspective, we provide a unified characterization of the asymptotic efficiency of the resulting estimators.

\subsection{Related work}

\paragraph{Scoring rules}
Proper scoring rules provide a framework for evaluating the quality of probability distributions \citep{Gneiting2007}. 
The seminal work of \citet{Hyvarinen2005} proposed score matching (Hyvärinen scoring rule) for the estimation of unnormalized models on $\R^p$.
Various extensions of score matching to more general domains have been developed \citep[e.g.,][]{Dawid2005,Hyvarinen2007,Lyu2009,Yu2019,Mardia2016,Liu2022,Williams2022,Scealy2023}.
To analyze the asymptotic efficiency of score matching estimators, \citet{Koehler2023} and \citet{Qin2024} exploited generators of diffusion processes associated with score matching.
Closely related to this paper is \citet{Lyu2009}, in which score matching was generalized by replacing the gradient operator with a general linear operator $\L$.
Another closely related work is \citet{Gutmann2011}, in which a unified framework for estimation of unnormalized models was developed based on Bregman divergences.
Our framework, which incorporates both a general linear operator $\L$ and a Bregman-type divergence generated by a concave function $h$, reveals the connection between the resulting scoring rules and Stein's method.

\citet{Parry2012} generalized score matching to local scoring rules and characterized their propriety.
\citet{Ehm2012} characterized the propriety of local scoring rules of order two.
\citet{Takasu2018} proposed proper local scoring rules on unit spheres.
\citet{Dawid2012} introduced local scoring rules on discrete spaces that include pseudolikelihood \citep{Besag1975} and ratio matching \citep{Hyvarinen2007} as special cases.
\citet{Kanamori2017} investigated strictly proper local scoring rules on discrete sample spaces through their relationship with Bregman divergences.
Our class of proper scoring rules recovers local scoring rules on both continuous and discrete sample spaces through appropriate choices of $\L$.
\citet{Marlin2011} studied the asymptotic relative efficiency of Besag's pseudolikelihood and ratio matching.
\citet{Dawid2016} discussed parametric estimation based on scoring rules.
\citet{Waghmare2026} provide a comprehensive review of recent developments in scoring rules.

An alternative approach is noise contrastive estimation (NCE), proposed by \citet{Gutmann2010}, which estimates the normalizing constant rather than employing a scoring rule that does not depend on it.
\citet{Matsuda2019} extended NCE to finite mixtures of unnormalized models.
\citet{Fujii2026} generalized NCE using Bregman divergences.
\citet{Sonobe2026} developed a fully Bayesian extension of the contrastive approach, thereby enabling posterior uncertainty quantification.
\citet{Uehara2020} developed a framework for asymptotically efficient estimation without the normalizing constant by utilizing a nonparametric density estimator.

\paragraph{Stein's method}
Stein's method introduced by \citet{Stein1972} was originally developed for normal approximation, but it now plays an important role in the estimation of unnormalized models \citep{Anastasiou2023,Liu2026}.
\citet{Barp2019} introduced minimum Stein discrepancy estimation and showed that score matching is recovered as a special case using the Langevin--Stein operator.
\citet{Grathwohl2020} used neural networks to parameterize the test functions in Stein discrepancies.
\citet{Ebner2025} proposed Stein's method of moments (SMoM), which yields a wide class of estimators that do not depend on the normalizing constant.
By choosing appropriate Stein operators, SMoM can be extended to more general domains \citep{Fischer2025discrete,Fischer2025truncated,Fischer2025graph,Fischer2026}.
\citet{Kume2026} showed that weighted score matching is an instance of SMoM.
\citet{Nagai2026} characterized the asymptotic efficiency of the score matching estimator using the geometry of SMoM.
Extending this geometric approach to our class of proper scoring rules, we show that the resulting estimators are asymptotically efficient if and only if the corresponding Stein score functions and the Fisher score functions span the same space.

For the construction of Stein operators, \citet{Gorham2015} and \citet{Gorham2019} proposed a generator approach that exploits the generators of Markov processes.
\citet{Ley2013} and \citet{Mijoule2023} introduced a density approach.
\citet{Ley2016spl,Ley2016bjps} proposed a parametric approach.
\citet{Yang2018} characterized Stein operators on discrete spaces.
\citet{Eguchi2026} derived the $\gamma$-Stein operator from the $\gamma$-divergence.
Our construction of Stein operators is based on the estimating equations, or equivalently on the first-order optimality conditions, associated with proper scoring rules.

\subsection{Organization}
The rest of this paper is organized as follows.
In Section~\ref{sec:preliminaries}, we introduce notation and review scoring rules and Stein's method.
In Section~\ref{sec:mainresults}, we introduce our class of proper scoring rules and the induced Stein operators.
In Section~\ref{sec:asymptoticefficiency}, we provide a unified characterization of the asymptotic efficiency of estimators based on our class of proper scoring rules.
In Section~\ref{sec:examples}, we present examples.
In the Supplementary Material, we state regularity conditions, provide further examples, and present proofs omitted from the main text.

\section{Preliminaries}
\label{sec:preliminaries}

In this section, we introduce notations and review scoring rules and Stein's method.

\subsection{Notation}

Let $(\X,\mathscr{B},\mathrm{d}x)$ be a measure space.
Let the parameter space $\Theta\subset\R^d$ be an open set,
and let $p_{\theta}$ be a probability density on $\X$ such that $p_{\theta}(x)>0$ and $\theta\mapsto p_{\theta}(x)$ is sufficiently smooth.
The parametric model $\P$ is defined by $\P\coloneqq\{p_{\theta}\mid \theta\in\Theta\}$.
We assume that $\P$ is identifiable, that is, $p_{\theta}=p_{\theta'}$ if and only if $\theta=\theta'$.
For an estimator $\hat\theta$ satisfying $\sqrt{n}(\hat\theta-\theta^\star)\warrow N(0,V)$, we set $\AVar[\hat\theta]\coloneqq V$.

The standard inner product and the induced norm on $\R^m$ are denoted by $\ang{\,\cdot\,,\,\cdot\,}$ and $\norm{\,\cdot\,}$, respectively.
The gradient operator on $\Theta$ is denoted by $\nabla_{\theta}$.
Let $\partial_{\theta_j}$ denote the partial differential operator $\partial/\partial\theta_{j}$.

Let $\Meas(\X;\R^m)$ be the space of all measurable functions from $\X$ to $\R^m$.
For $r\in\mathbb{N}$, define
\[
    L^r(p_{\theta})\coloneqq\{f\in\Meas(\X;\R)\mid\E_{\theta}[|f(X)|^r]<\infty\},\quad 
    L^r(\P)\coloneqq\bigcap_{\theta\in\Theta}L^r(p_{\theta}),
\]
where functions equal almost everywhere are identified in the usual manner.

\subsection{Scoring rules}

A scoring rule is a loss function $S\colon\X\times\P\to\R$ such that $(x\mapsto S(x,p_{\theta}))\in L^1(\P)$ for all $\theta\in\Theta$.
The scoring rule $S$ is said to be proper if for any (true) parameter $\theta^\star\in\Theta$, $\theta\mapsto\E_{\theta^\star}[S(X,p_{\theta})]$ is minimized at $\theta=\theta^\star$.
$S$ is said to be strictly proper if the minimizer is unique.
Propriety justifies the use of a scoring rule $S$ as a loss function.
A proper scoring rule $S$ induces the statistical divergence 
\[
D(p_{\theta^\star}\,\|\,p_{\theta})\coloneqq\E_{\theta^\star}[S(X,p_{\theta})]-\E_{\theta^\star}[S(X,p_{\theta^\star})].
\]
$D(p_{\theta^\star}\,\|\,p_{\theta})\geq0$ whenever $S$ is proper.
If $S$ is strictly proper, $D(p_{\theta^\star}\,\|\,p_{\theta})=0$ if and only if $\theta=\theta^\star$.

Let $X_1,\dots,X_n$ be i.i.d.~random variables from $p_{\theta^\star}$.
The estimator $\hat\theta$ based on a proper scoring rule $S$ is defined by
\begin{equation}
\label{eq:scoringruleestimator}    
\hat\theta \in \underset{\theta\in\Theta}{\arg\min}\,\frac{1}{n}\sum_{i=1}^{n}S(X_i,p_{\theta}).
\end{equation}
The maximum likelihood estimator $\hat\theta_{\mathrm{MLE}}$ corresponds to the log score $S(x,p_{\theta})\coloneqq-\log p_{\theta}(x)$.

\subsection{Stein's method}

For $\theta\in\Theta$, let $\A_{\theta}$ be a linear operator on $L^2(p_{\theta})$ with the domain $\operatorname{Dom}\A_{\theta}$.
Let $\F_{\theta}\subset\{f\in\operatorname{Dom}\A_{\theta}\mid \A_{\theta}f\in L^1(\P)\}$ be a class of test functions.
The pair $(\A_{\theta},\F_{\theta})$ is said to be a Stein pair if for any $\theta^\star\in\Theta$,
\[
\E_{\theta^\star}[\A_{\theta}f(X)]=0, \quad \forall f\in\F_{\theta} \iff \theta=\theta^\star.
\]
For a Stein pair $(\A_{\theta},\F_{\theta})$, $\A_{\theta}$ is called a Stein operator and $\F_{\theta}$ is called a Stein class.
We provide some examples of Stein operators in Section~\ref{sec:examples}.

\section{Main results}
\label{sec:mainresults}

In this section, we introduce the class of proper scoring rules for unnormalized models and the Stein operators induced by these scoring rules.

\subsection{Proper scoring rules for unnormalized models}

We construct scoring rules as follows.
Let $\mathcal{Y}\subset\R^m$ be an open convex set and $h\colon\X\times\mathcal{Y}\to\R$ be a measurable function such that $y\mapsto h(x,y)$ is sufficiently smooth.
The gradient and Hessian of $h$ with respect to $y$ are denoted by $h_{y}(x,y)$ and $h_{yy}(x,y)$, respectively.
Here, the second argument of $h_y$ and $h_{yy}$ specifies the evaluation point of the derivatives with respect to $y$.
Let $\L\colon\operatorname{Dom}\L\subset \Meas(\X;\R) \to \Meas(\X;\R^m)$ be a linear operator satisfying $p_{\theta}\in\operatorname{Dom}\L$ and $\L p_{\theta}(x)/p_{\theta}(x)\in\mathcal{Y}$ for all $x\in\X$ and $\theta\in\Theta$.
Let $\L^\ast\colon \operatorname{Dom}\L^\ast\subset\Meas(\X;\R^m)\to \Meas(\X;\R)$ be a linear operator that is adjoint to $\L$ in the following sense: for every $f\in\operatorname{Dom}\L$ and $V\in\operatorname{Dom}\L^\ast$,
\begin{equation}
    \label{eq:adjoint}
    \int_{\X} f(x)\L^\ast V(x) \d x = \int_{\X}\ang*{V(x),\L f(x)}\d x,
\end{equation}
whenever $f\L^\ast V$ and $\ang{V(\,\cdot\,),\L f(\,\cdot\,)}$ are integrable.

We assume the following identifiability condition on $\L$.

\begin{assumption}
\label{assumption:L-identifiability}
For all $\theta,\theta^\star\in\Theta$,
the equation
$\L p_{\theta}/p_{\theta}=\L p_{\theta^\star}/p_{\theta^\star}$ holds if and only if $\theta=\theta^\star$.
\end{assumption}

We impose regularity conditions on $\L$ and $h$ detailed in Appendix~\ref{sec:RegularityConditions}, ensuring that all relevant expectations are finite and the functions involved lie in the appropriate operator domains.
Under the regularity conditions, the scoring rule $S\colon\X\times\P\to\R$ is defined by
\begin{align}
\label{eq:scoringrule}
    &S(x,p_{\theta})\coloneqq h\left(x,\frac{\L p_{\theta}(x)}{p_{\theta}(x)}\right) +\L^\ast h_{y}\left(\,\cdot\,,\frac{\L p_{\theta}(\,\cdot\,)}{p_{\theta}(\,\cdot\,)}\right)(x)-\ang*{h_{y}\left(x,\frac{\L p_{\theta}(x)}{p_{\theta}(x)}\right),\frac{\L p_{\theta}(x)}{p_{\theta}(x)}}.
\end{align}
In general, $S$ does not require computation of the normalizing constant, since $\L p_\theta/p_\theta$ is independent of it.
The exception is when $\L$ itself generates an additional term involving the normalizing constants; see Sections~\ref{sec:ExamplesNCE} and~\ref{sec:ExamplesMLE}.

If $y\mapsto h(x,y)$ is concave, the scoring rule $S$ induces the following Bregman-type divergence:
\begin{align}
\label{eq:bregman}
D(p_{\theta^\star}\,\|\,p_{\theta})
&\coloneqq\E_{\theta^\star}[S(X,p_{\theta})-S(X,p_{\theta^\star})]\notag\\
&=\E_{\theta^\star}\left[
h\left(X,\frac{\L p_{\theta}(X)}{p_{\theta}(X)}\right)-h\left(X,\frac{\L p_{\theta^\star}(X)}{p_{\theta^\star}(X)}\right)\right.\notag\\
&\left.\phantom{=}-\ang*{h_{y}\left(X,\frac{\L p_{\theta}(X)}{p_{\theta}(X)}\right),\frac{\L p_{\theta}(X)}{p_{\theta}(X)}-\frac{\L p_{\theta^\star}(X)}{p_{\theta^\star}(X)}}
\right].
\end{align}    

The following theorem shows that a linear operator $\L$ and a concave function $h$ yield a proper scoring rule.

\begin{theorem}
    \label{thm:propriety}
    Suppose $y\mapsto h(x,y)$ is concave.
    Then, the scoring rule $S$ defined by \eqref{eq:scoringrule} is proper.
    Furthermore, if $y\mapsto h(x,y)$ is strictly concave, and if $\L$ satisfies Assumption~\ref{assumption:L-identifiability}, $S$ is strictly proper.
\end{theorem}

\begin{proof}
    By the concavity of $h$, $D(p_{\theta^\star}\,\|\,p_{\theta})$ defined by \eqref{eq:bregman} is nonnegative for any $\theta,\theta^\star\in\Theta$.
    Thus, $S$ is a proper scoring rule. 
    Furthermore, if $h$ is strictly concave, $D(p_{\theta^\star}\,\|\,p_{\theta})=0$ if and only if $\L p_{\theta}/p_{\theta}=\L p_{\theta^\star}/p_{\theta^\star}$.
    Under Assumption~\ref{assumption:L-identifiability}, this is equivalent to $\theta=\theta^\star$.
    Therefore, $S$ is strictly proper, which completes the proof.
\end{proof}

\begin{remark}[Comparison with \citet{Lyu2009}]
     Generalized score matching \citep{Lyu2009} is the proper scoring rule given by
    \[
        S(x,p_{\theta})\coloneqq \frac12\norm*{\frac{\L p_{\theta}(x)}{p_{\theta}(x)}}^2 -\L^\ast\left[\frac{\L p_{\theta}}{p_{\theta}}\right](x).
    \]
    This scoring rule is recovered by choosing $h(x,y)\coloneqq-\norm{y}^2/2$.
    Allowing a general concave function $h$ is essential for the scope of our framework: together with appropriate choices of $\L$, it encompasses important examples; see Section~\ref{sec:examples}.
    Moreover, the curvature $h_{yy}$ determines the geometry of the associated divergence and, consequently, the induced Stein operator.
\end{remark}

\begin{remark}[Comparison with \citet{Takasu2018}]
    \citet{Takasu2018} proposed proper local scoring rules on the unit sphere $\mathbb{S}^{p-1}$.
    Their construction is recovered by choosing the Riemannian gradient $\L\coloneqq\nabla_{\mathbb{S}^{p-1}}$.
\end{remark}

\begin{remark}[Comparison with \citet{Gutmann2011}]
    For estimation of unnormalized models, \citet{Gutmann2011} provided a unified framework based on Bregman divergences.
    Although their framework encompasses the divergence defined in \eqref{eq:bregman}, our focus is different.
    We study Bregman divergences that are empirically estimable through proper scoring rules. 
    This representation naturally induces an associated Stein operator and enables a unified analysis of the asymptotic efficiency of the resulting estimators.
\end{remark}

\subsection{Stein operators induced by proper scoring rules}

For each $\theta\in\Theta$, the linear operator $\A_{\theta}$ on $L^2(p_{\theta})$ is defined as follows.
\begin{align}
\label{eq:SteinOperator}    
\A_{\theta}f(x)
\coloneqq \L^\ast\left[ h_{yy}\left(\,\cdot\,,\frac{\L p_{\theta}(\,\cdot\,)}{p_{\theta}(\,\cdot\,)}\right)\D_{\theta}f\right](x)-\ang*{h_{yy}\left(x,\frac{\L p_{\theta}(x)}{p_{\theta}(x)}\right)\D_{\theta}f(x),\frac{\L p_{\theta}(x)}{p_{\theta}(x)}},
\end{align}
where $\D_{\theta}f$ is defined by
\[
\D_{\theta}f(x)\coloneqq \frac{\L[fp_{\theta}](x)-f(x)\L p_{\theta}(x)}{p_{\theta}(x)}.
\]

The domain $\operatorname{Dom}\A_{\theta}\subset L^2(p_{\theta})$ and its subset $\F_{\theta}\subset\operatorname{Dom}\A_{\theta}$
are chosen to satisfy appropriate integrability conditions and domain requirements for $\L$ and $\L^\ast$.
The details are given in Appendix~\ref{sec:RegularityConditions}.

To derive $\A_{\theta}$ from the proper scoring rule $S$ defined by \eqref{eq:scoringrule}, we make the following commutation assumption.
\begin{assumption}
    \label{assumption:Commutation}
    For $j=1,\dots,d$,
    \begin{align*}    
        \partial_{\theta_j}\L p_{\theta}&=\L\partial_{\theta_j}p_{\theta},\\
        \partial_{\theta_j}\L^\ast\left[h_{y}\left(\,\cdot\,,\frac{\L p_{\theta}(\,\cdot\,)}{p_{\theta}(\,\cdot\,)}\right)\right]&=\L^\ast\left[\partial_{\theta_j}h_{y}\left(\,\cdot\,,\frac{\L p_{\theta}(\,\cdot\,)}{p_{\theta}(\,\cdot\,)}\right)\right].
    \end{align*}
\end{assumption}

The following theorem states that the linear operator $\A_{\theta}$ is naturally induced by the proper scoring rule $S$ and that Stein's identity holds.

\begin{theorem}
    \label{thm:ScoringRuleStein}
    Let $S$ be the scoring rule defined by \eqref{eq:scoringrule} and $\A_{\theta}$ be the linear operator defined by \eqref{eq:SteinOperator}.
    \begin{enumerate}
        \item Suppose $\partial_{\theta_j}\log p_{\theta}\in\operatorname{Dom}\A_{\theta}$ for $j=1,\dots,d$.
        Then, under Assumption~\ref{assumption:Commutation}, we have
        \[
        \partial_{\theta_j}S(x,p_{\theta})=\A_{\theta}\partial_{\theta_j}\log p_{\theta}(x),\quad j=1,\dots,d.
        \]
        \item For any $f\in\operatorname{Dom}\A_{\theta}$, we have $\E_{\theta}[\A_{\theta}f(X)]=0$.
    \end{enumerate}
\end{theorem}
\begin{proof}
    $(1)$: Observe that
    \[
    \partial_{\theta_j}\frac{\L p_{\theta}(x)}{p_{\theta}(x)}=\D_{\theta}\partial_{\theta_j}\log p_{\theta}(x),\quad j=1,\dots,d.
    \]
    Thus, for $j=1,\dots,d$, we have
    \begin{align*}
     \partial_{\theta_j}S(x,p_{\theta})
        &=\ang*{h_{y}\left(x,\frac{\L p_{\theta}(x)}{p_{\theta}(x)}\right),\D_{\theta}\partial_{\theta_j}\log p_{\theta}(x)}
        +\L^\ast\left[ h_{yy}\left(\,\cdot\,,\frac{\L p_{\theta}(\,\cdot\,)}{p_{\theta}(\,\cdot\,)}\right)\D_{\theta}\partial_{\theta_j}\log p_{\theta}\right](x)\\
        &\phantom{=}-\ang*{h_{yy}\left(x,\frac{\L p_{\theta}(x)}{p_{\theta}(x)}\right)\D_{\theta}\partial_{\theta_j}\log p_{\theta}(x),\frac{\L p_{\theta}(x)}{p_{\theta}(x)}}
        -\ang*{h_{y}\left(x,\frac{\L p_{\theta}(x)}{p_{\theta}(x)}\right),\D_{\theta}\partial_{\theta_j}\log p_{\theta}(x)}\\
        &=\A_{\theta}\partial_{\theta_j}\log p_{\theta}(x),
    \end{align*}
    which proves $(1)$.
    $(2)$: This follows directly from the adjoint relation \eqref{eq:adjoint}.
\end{proof}

\begin{remark}[Connection to SMoM]
    Let $X_1,\dots,X_n$ be i.i.d. random variables from $p_{\theta^\star}$.
    A Stein's method of moments (SMoM) estimator with respect to the test functions $f_{\theta,1},\dots,f_{\theta,d}\in\operatorname{Dom}\A_{\theta}$ is defined as a solution to the following estimating equations \citep{Ebner2025}:
    \[
    \sum_{i=1}^{n}\A_{\theta}f_{\theta,j}(X_i)=0,\quad j=1,\dots,d.
    \]
    Part~(1) of Theorem~\ref{thm:ScoringRuleStein} shows that the first-order estimating equations associated with the estimator defined by \eqref{eq:scoringruleestimator}, with $S$ given by \eqref{eq:scoringrule}, coincide with those of an SMoM estimator with test functions $f_{\theta,j}\coloneqq\partial_{\theta_j}\log p_{\theta}$, $j=1,\dots,d$.
    This relation provides a canonical choice of test functions \citep{Nagai2026}.
\end{remark}

\begin{remark}[Comparison with \citet{Eguchi2026}]
    \citet{Eguchi2026} derived the $\gamma$-Stein operator by taking the first variation of the $\gamma$-divergence along infinitesimal transports induced by a geometry on the sample space.
    In contrast, our construction is based on the divergence geometry induced by a scoring rule. 
    Consequently, when applied to the $\gamma$-divergence (pseudo-spherical scoring rule), our construction yields a Stein operator distinct from the $\gamma$-Stein operator of \citet{Eguchi2026}; see Appendix~\ref{sec:ExamplesLocalScoringRuleDiscrete}.
\end{remark}

The following proposition gives a sufficient condition for $(\A_{\theta},\F_{\theta})$ to be a Stein pair under a strong integrability condition.
In Section~\ref{sec:examples}, we relax this condition for specific choices of $\L$.

\begin{proposition}
    \label{prp:SteinFormContraction}
    Suppose that $p_{\theta^\star}/p_{\theta}\in\F_{\theta}$ for all $\theta^\star,\theta\in\Theta$ 
    and that $h_{yy}(x,y)$ is negative definite.
    Then, under Assumption~\ref{assumption:L-identifiability}, $(\A_{\theta},\F_{\theta})$ is a Stein pair.
\end{proposition}
\begin{proof}
    Fix an arbitrary $\theta^\star\in\Theta$.
    We show that for $f_{\theta}\coloneqq p_{\theta^\star}/p_{\theta}\in\F_{\theta}$,
    $\E_{\theta^\star}[\A_{\theta}f_{\theta}(X)]=0$ implies $\theta=\theta^\star$.
    Observe that
    \[
        \D_{\theta}\left[\frac{p_{\theta^{\star}}}{p_{\theta}}\right]
        =\frac{p_{\theta^{\star}}}{p_{\theta}}\left(\frac{\L p_{\theta^\star}}{p_{\theta^\star}}-\frac{\L p_{\theta}}{p_{\theta}}\right).
    \]
    Thus, we have
    \begin{align*}
    \E_{\theta^\star}\left[\A_{\theta}f_{\theta}(X)\right]
    =\E_{\theta^\star}\left[\frac{p_{\theta^\star}(X)}{p_{\theta}(X)}\ang*{h_{yy}\left(X,\frac{\L p_{\theta}(X)}{p_{\theta}(X)}\right)\left(\frac{\L p_{\theta^\star}(X)}{p_{\theta^\star}(X)}-\frac{\L p_{\theta}(X)}{p_{\theta}(X)}\right),\frac{\L p_{\theta^\star}(X)}{p_{\theta^\star}(X)}-\frac{\L p_{\theta}(X)}{p_{\theta}(X)}}\right]
    \end{align*}
    Since $h_{yy}(x,\L p_{\theta}(x)/p_{\theta}(x))$ is negative definite, 
    \[
        \E_{\theta^\star}\left[\A_{\theta}f_{\theta}(X)\right]=0
        \iff \frac{\L p_{\theta^\star}}{p_{\theta^\star}}=\frac{\L p_{\theta}}{p_{\theta}}.
    \]
    Under Assumption~\ref{assumption:L-identifiability}, $\L p_{\theta^\star}/p_{\theta^\star}=\L p_{\theta}/p_{\theta}$ is equivalent to $\theta=\theta^\star$.
    Together with part~(2) of Theorem~\ref{thm:ScoringRuleStein}, this proves that $(\A_{\theta},\F_{\theta})$ is a Stein pair.
\end{proof}

We provide an example in which the strong integrability condition in Proposition~\ref{prp:SteinFormContraction} fails.
\begin{example}[Normal distribution]
    \label{exm:SteinFormContractionExampleNormal}
    Consider the family of zero-mean normal distributions $\P\coloneqq\{p_{\theta}(x)\coloneqq\exp(-x^2/2\theta^2)/\sqrt{2\pi\theta^2}\mid \theta>0\}$.
    We choose $\L\coloneqq\nabla_x,\L^\ast\coloneq-\div_x$, and $h(x,y)\coloneqq-\norm{y}^2/2$, which yields score matching.
    Observe that
    \[
    \E_{\theta^\star}\left[\left|\A_{\theta}\left[\frac{p_{\theta^\star}}{p_{\theta}}\right](X)\right|\right]
    \propto\int_{0}^{\infty}\left|1-\frac{x^2}{\theta^{\star2}}\right|\exp\left\{\left(\frac{1}{2\theta^2}-\frac{1}{\theta^{\star2}}\right)x^2\right\}\d x.
    \]
    If $1/(2\theta^2)-1/\theta^{\star2}\geq0$, this expectation is not finite.
    Hence, $p_{\theta^\star}/p_{\theta}\notin\F_{\theta}$ for such choices of $\theta$ and $\theta^\star$, so the integrability condition in Proposition~\ref{prp:SteinFormContraction} fails.
    This issue can be avoided by using Proposition~\ref{prp:SteinFormContractionScoreMatching}.
    In this case, we can take $\Psi(u)\coloneqq\log u$.
    Indeed, for all $\theta,\theta^\star>0$,
    \[
    \E_{\theta^\star}\left[\left|\A_{\theta}\log\frac{p_{\theta^\star}}{p_{\theta}}(X)\right|\right]
    =\left|\frac{1}{\theta^2}-\frac{1}{\theta^{\star2}}\right|\E_{\theta^\star}\left[\left|1-\frac{X^2}{\theta^2}\right|\right]<\infty.
    \]
\end{example}

\section{Unified characterization of asymptotic efficiency}
\label{sec:asymptoticefficiency}

In this section, we provide the unified characterization of the asymptotic efficiency of the estimator $\hat\theta$ defined in \eqref{eq:scoringruleestimator} with the scoring rule $S$ defined in \eqref{eq:scoringrule}.
Throughout this section, we assume that \(y\mapsto h(x,y)\) is concave and that Assumptions~\ref{assumption:L-identifiability} and \ref{assumption:Commutation} hold.

For each $\theta\in\Theta$, the bilinear form $\EE_{\theta}\colon \operatorname{Dom}\A_{\theta}\times\operatorname{Dom}\A_{\theta}\to\R$ is defined by
\begin{align}
    \label{eq:SteinForm}
    \EE_{\theta}(f,g)\coloneqq-\E_{\theta}\left[\ang*{h_{yy}\left(X,\frac{\L p_{\theta}(X)}{p_{\theta}(X)}\right)\D_{\theta}f(X),\D_{\theta}g(X)} \right].
\end{align}

We refer to $\EE_{\theta}$ as the \textit{Dirichlet form},
since it coincides with the Dirichlet form associated with $\A_{\theta}$ whenever $\A_{\theta}$ is the generator of a Markov process with invariant distribution $p_{\theta}$.
The following lemma shows the connection between $\A_{\theta}$ and $\EE_{\theta}$.
The proof is given in Appendix~\ref{sec:proof_AsymptoticEfficiency}.

\begin{lemma}
    For all $f,g\in\operatorname{Dom}\A_{\theta}$, we have
    \label{lem:SteinFormSymmetry}
    \begin{align*}
        \EE_{\theta}(f,g)
        =-\E_{\theta}[f(X)\A_{\theta}g(X)]=-\E_{\theta}[g(X)\A_{\theta}f(X)].
    \end{align*}
\end{lemma}

\begin{remark}[Connection to the geometry of proper scoring rules]
    The Dirichlet form $\EE_{\theta}$ underlies the local geometry induced by the scoring rule $S$ defined in \eqref{eq:scoringrule}, or equivalently by the associated Bregman divergence defined in \eqref{eq:bregman} \citep{Eguchi1992,Dawid2007}.
    In particular, the induced metric tensor on the parametric model $\P$ is represented by $G(\theta)_{jk}=\EE_{\theta}(\partial_{\theta_j}\log p_{\theta},\partial_{\theta_k}\log p_{\theta})$.
\end{remark}

The Stein score functions, which play a key role in the characterization of asymptotic efficiency, are defined as follows.
Their existence and uniqueness are discussed in Appendix~\ref{sec:ExistenceUniquenessSteinScore}.

\begin{definition}[Stein score]
    \label{def:SteinScore}
    For $j=1,\dots,d$, a Stein score function $\psi_{\theta,j}\in\operatorname{Dom}\A_{\theta}$ is a function satisfying
    \[
    \A_{\theta}\psi_{\theta,j}=-\partial_{\theta_j}\log p_{\theta},\enspace \E_{\theta}[\psi_{\theta,j}(X)]=0.
    \]
\end{definition}

We impose the following kernel condition on $\EE_{\theta}$.

\begin{assumption}
    \label{assumption:SteinFormKernel}
    For all $\theta\in\Theta$ and $f\in\operatorname{Dom}\A_{\theta}$, the equation $\EE_{\theta}(f,f)=0$ holds only if $f$ is constant.
\end{assumption}

We now present the unified characterization of the asymptotic efficiency of $\hat\theta$.
We assume the regularity conditions on the scoring rule $S$ and the model $\P$ detailed in Appendix~\ref{sec:RegularityConditions}.
Our main assumptions are the consistency and asymptotic linearity of $\hat\theta$ and $\hat\theta_\mathrm{MLE}$, which can be established using standard arguments for M-estimators \citep[e.g.,][]{VanderVaart1998}. 

The following theorem shows that $\hat\theta$ is asymptotically efficient if and only if the Fisher score functions span the same space as the Stein score functions.
The proof is given in Appendix~\ref{sec:proof_AsymptoticEfficiency}.

\begin{theorem}
    \label{thm:AsymptoticEfficiency}
    Suppose that the regularity conditions R1--R5 in Appendix~\ref{sec:RegularityConditions} hold and that Stein score functions $\psi_{\theta^\star,1},\dots,\psi_{\theta^\star,d}$ exist.
    Let $\hat\theta$ be the estimator defined in \eqref{eq:scoringruleestimator} with the proper scoring rule $S$ defined in \eqref{eq:scoringrule}.
    Under Assumption~\ref{assumption:SteinFormKernel}, the following are equivalent:
    \begin{enumerate}
        \item $\hat\theta$ is asymptotically efficient, that is, $\AVar[\hat\theta]=\AVar[\hat\theta_\mathrm{MLE}]$;
        \item there exists a nonsingular matrix $\Lambda(\theta^\star)\in\R^{d\times d}$ such that
        $\partial_{\theta_j}\log p_{\theta^\star}=\sum_{k=1}^{d}\Lambda(\theta^\star)_{jk}\psi_{\theta^\star,k}$ for $j=1,\dots,d$.
    \end{enumerate}
\end{theorem}

\begin{remark}[Comparison with \citet{Nagai2026}]
    \citet{Nagai2026} established the score matching counterpart of Theorem~\ref{thm:AsymptoticEfficiency}.
    We generalize their result to our class of proper scoring rules.
    Our proof follows their strategy and relies on the canonical decomposition of the estimator $\hat\theta$.
    This decomposition is derived from the orthogonal decomposition of the Stein score functions with respect to the Dirichlet form $\EE_{\theta^\star}$.
\end{remark}

\section{Examples}
\label{sec:examples}

In this section, we illustrate our framework through several examples.
We first summarize common sufficient conditions for Assumptions~\ref{assumption:L-identifiability} and \ref{assumption:SteinFormKernel}.
For every operator $\L$ considered in this section, Assumption~\ref{assumption:L-identifiability} is satisfied, and Assumption~\ref{assumption:SteinFormKernel} is also satisfied whenever $h_{yy}(x,y)$ is negative definite; see Appendix~\ref{sec:SuffCondForAsm} and Appendix~\ref{sec:ProofExample}.
Assumption~\ref{assumption:Commutation} is imposed separately whenever needed.

\subsection{Gradient operator and score matching}
\label{sec:ExamplesScoreMatching}

Let $\X\coloneqq\R^p$ and suppose $(x,\theta)\mapsto p_{\theta}(x)$ is smooth on $\X\times\Theta$.
We choose $\L$ as the Euclidean gradient operator: $\L\coloneqq\nabla_x$.
$\L^\ast$ is given by
$\L^\ast V(x)\coloneqq-\div_xV(x)=-\sum_{j=1}^{p}{\partial V_j(x)}/{\partial x_j}$.
The adjoint relation between $\L$ and $\L^\ast$ in \eqref{eq:adjoint} is justified by integration by parts under an appropriate boundary condition for $\P$.
This setting corresponds to the class of local scoring rules of order two \citep{Parry2012,Ehm2012}, and score matching \citep{Hyvarinen2005} is recovered by choosing $h(x,y)\coloneqq-\norm{y}^2/2$.
The scoring rule $S$, Stein operator $\A_{\theta}$, divergence $D(p_{\theta^\star}\,\|\,p_{\theta})$, and Dirichlet form $\EE_{\theta}$ corresponding to score matching are given as follows:
\begin{align*}
    S(x,p_{\theta})&=\Delta_x\log p_{\theta}(x)+\frac{1}{2}\norm{\nabla_x\log p_{\theta}(x)}^2,\\
    \A_{\theta}f(x)&=\Delta_x f(x)+\ang{\nabla_x f(x),\nabla_x \log p_{\theta}(x)},\\
    D(p_{\theta^\star}\,\|\,p_{\theta})&=\frac12\E_{\theta^\star}\left[\norm*{\nabla_x\log\frac{p_{\theta^\star}(X)}{p_{\theta}(X)}}^2\right],\\
    \EE_{\theta}(f,g)&=\E_{\theta}[\ang{\nabla_x f(X),\nabla_x g(X)}],
\end{align*}
where $\Delta_x$ is the Laplacian operator defined by $\Delta_x\coloneqq\div_x\nabla_x=\sum_{j=1}^{p}\partial^2/\partial x_j^2$.

\begin{remark}[Comparison with score matching on Riemannian manifolds]
   When $\X\subset\R^p$ is an oriented and connected Riemannian manifold with corners,
   we can choose $\L$ as the Riemannian gradient, which recovers score matching on Riemannian manifolds \citep{Dawid2005,Mardia2016} and its variants \citep{Takasu2018,Williams2022,Scealy2023}.
\end{remark}

The following proposition states an alternative sufficient condition for $(\A_{\theta},\F_{\theta})$ to be a Stein pair.
The proof is given in Appendix~\ref{sec:ProofExampleScoreMatching}.

\begin{proposition}
    \label{prp:SteinFormContractionScoreMatching}
    Let $\L\coloneqq\nabla_x,\L^\ast\coloneqq-\div_x$ and suppose $h_{yy}(x,y)$ is negative definite.
    Suppose further that there exists a smooth function $\Psi\colon(0,\infty)\to\R$ such that $\Psi'>0$ and $\Psi(p_{\theta^\star}(\,\cdot\,)/p_{\theta}(\,\cdot\,))\in\F_{\theta}$ for all $\theta^\star,\theta\in\Theta$.
    Then $(\A_{\theta},\F_{\theta})$ is a Stein pair.
\end{proposition}

\begin{remark}[Connection to entropy dissipation]
    In the setting of Proposition~\ref{prp:SteinFormContractionScoreMatching},
    let $\Phi\colon(0,\infty)\to\R$ be a smooth convex function such that $\Phi'=\Psi$.
    Suppose $\A_{\theta}$ is the generator of a Markov process with invariant distribution $p_{\theta}$, and let $(P_t^\theta)_{t\geq0}$ be the associated Markov semigroup.
    For a measurable function $f\colon\X\to(0,\infty)$ satisfying the required integrability conditions, the $\Phi$-entropy $\mathrm{Ent}_{\theta}^\Phi(f)$ is defined by
    \[
    \mathrm{Ent}_{\theta}^\Phi(f)\coloneqq\E_{\theta}[\Phi(f(X))]-\Phi(\E_{\theta}[f(X)]).
    \]
     Under suitable regularity conditions, a de Bruijn-type identity \citep[Proposition 1]{Chafai2004}, together with Lemma~\ref{lem:SteinFormSymmetry}, yields
    \[
        \left.\frac{\mathrm{d}}{\mathrm{d}t}\,\mathrm{Ent}_{\theta}^{\Phi}\left( P_t^\theta\left[\frac{p_{\theta^\star}}{p_{\theta}}\right]\right)\right|_{t=0+}
        =\E_{\theta^\star}\left[r_{\theta}(X)\right],
    \]
    where $r_{\theta}(x)\coloneqq\A_{\theta}\Phi'({p_{\theta^\star}(\,\cdot\,)}/{p_{\theta}(\,\cdot\,)})(x)$.
    Hence, Proposition~\ref{prp:SteinFormContractionScoreMatching} implies that the initial $\Phi$-entropy dissipation rate vanishes if and only if $\theta=\theta^\star$.
\end{remark}

The following proposition provides a class of models $\P$ satisfying the condition in part (2) of Theorem~\ref{thm:AsymptoticEfficiency}.
The proof is given in Appendix~\ref{sec:ProofExampleScoreMatching}.

\begin{proposition}
    \label{prp:AsymptoticEfficiencyExampleScoreMatching}
    Let $\L\coloneqq\nabla_x$, $\L^\ast\coloneqq-\div_x$, and let $p(x)>0$ be a probability density on $\R^p$ with zero mean such that $\log p$ is smooth and $\nabla_x^2\log p(x)$ is negative definite.
    Let $h(x,y)\coloneqq-(-\log p)^\ast(-y)$, where $(-\log p)^\ast$ is the Legendre transform of $-\log p$ defined by
    \[(-\log p)^\ast(u)\coloneqq\sup_{x\in\R^p}\{\ang{u,x}+\log p(x)\}.\]
    Consider the location model $p_{\theta}(x)\coloneqq p(x-\theta)$, where $\Theta=\R^p$.
    Then, the Stein score functions coincide with the Fisher score functions; that is, $\psi_{\theta,j}=\partial_{\theta_j}\log p_{\theta}$ for $j=1,\dots,p$.
\end{proposition}

\begin{remark}[Connection to Wasserstein geometry]
    Under the score matching choice
    \[
        \L\coloneqq\nabla_{x},\,
        \L^\ast\coloneqq-\div_x,\,
        h(x,y)\coloneqq -\frac12\norm{y}^2,
    \]
    the Stein score functions coincide with the Wasserstein score functions \citep{Li2023,Nishimori2025,Nagai2026} that play a key role in Wasserstein geometry.

    For the location model in Proposition~\ref{prp:AsymptoticEfficiencyExampleScoreMatching},
    the Wasserstein score functions are given by $\psi_{\theta,j}(x)=x_j-\theta_j$ for $j=1,\dots,p$, and hence do not encode the shape of the base density $p(x)$  \citep{Li2023,Amari2024,Fukushi2026}.
    In contrast, the choice of $h$ in Proposition~\ref{prp:AsymptoticEfficiencyExampleScoreMatching} incorporates the shape of $p(x)$ through the geometry induced by $h_{yy}$.
\end{remark}

\begin{remark}[Comparison with \cite{Koehler2023} and \cite{Qin2024}]
    \citet{Koehler2023} related the asymptotic efficiency of score matching to the Poincaré inequality for the Langevin diffusion. \citet{Qin2024} generalized this connection to a broader class of Markov processes associated with the generator $\A_\theta$ and diffusion score matching losses.
    Both results give an upper bound on the asymptotic variance of score matching.
    Our characterization gives a different perspective: the loss of asymptotic efficiency is quantified by the leakage of the Stein score functions from the space spanned by the Fisher score functions, as made explicit by the asymptotic variance decomposition in Appendix~\ref{sec:proof_AsymptoticEfficiency}.
\end{remark}

\subsection{Flip operator and ratio matching}
\label{sec:ExamplesRatioMatching}

Let $\X\coloneqq\{\pm1\}^p$ and
\begin{equation}
    \label{eq:FlipOperator}
    \L f(x)\coloneqq(f(x^{(1)}),\dots,f(x^{(p)}))^\top,
\end{equation}
where $x^{(j)}\coloneqq(x_1,\dots,-x_{j},\dots,x_p)^\top$.
$\L^\ast$ is given by
\begin{equation}
    \label{eq:FlipOperatorAdjoint}    
    \L^\ast V(x)\coloneqq\sum_{j=1}^{p}V_j(x^{(j)}).
\end{equation}
In this setting, ratio matching \citep{Hyvarinen2007} is recovered by choosing $h(x,y)\coloneqq\sum_{j=1}^{p}y_j/(4+4y_j)$.
The corresponding scoring rule $S$, Stein operator $\A_{\theta}$, divergence $D(p_{\theta^\star}\,\|\,p_{\theta})$, and Dirichlet form $\EE_{\theta}$ are given as follows:
\begin{align*}
    S(x,p_{\theta})&=
    \frac12\sum_{j=1}^{p}w_{\theta,j}(x)^2,\\
    \A_{\theta}f(x)&=
    \sum_{j=1}^{p}w_{\theta,j}(x)^3\frac{p_{\theta}(x)}{p_{\theta}(x^{(j)})}(f(x^{(j)})-f(x)),\\
    D(p_{\theta^\star}\,\|\,p_{\theta})&=\frac12\sum_{j=1}^{p}\E_{\theta^\star}\left[(w_{\theta^\star,j}(X)-w_{\theta,j}(X))^2\right],\\
    \EE_{\theta}(f,g)&=
    \frac12\sum_{j=1}^{p}\E_{\theta}\left[w_{\theta,j}(X)^3\frac{p_{\theta}(X)}{p_{\theta}(X^{(j)})}(f(X^{(j)})-f(X))(g(X^{(j)})-g(X))\right],
\end{align*}
where $w_{\theta,j}(x)\coloneqq p_{\theta}(x^{(j)})/(p_{\theta}(x)+p_{\theta}(x^{(j)}))$.

The following proposition gives an alternative sufficient condition for $(\A_{\theta},\F_{\theta})$ to be a Stein pair.
The proof is given in Appendix~\ref{sec:ProofExampleRatioMatching}.

\begin{proposition}
    \label{prp:SteinFormContractionRatioMatching}
    Let $\L$ and $\L^\ast$ be defined in \eqref{eq:FlipOperator} and \eqref{eq:FlipOperatorAdjoint}, respectively.
    Suppose $h_{yy}(x,y)$ is negative definite and diagonal.
    Suppose further that there exists a strictly increasing function $\Psi\colon(0,\infty)\to\R$ such that $\Psi(p_{\theta^\star}(\,\cdot\,)/p_{\theta}(\,\cdot\,))\in\F_{\theta}$ for all $\theta^\star,\theta\in\Theta$.
    Then $(\A_{\theta},\F_{\theta})$ is a Stein pair.
\end{proposition}

The following proposition provides a class of models $\P$ satisfying the condition in part (2) of Theorem~\ref{thm:AsymptoticEfficiency}.
Note that the choice of $h$ in Proposition~\ref{prp:AsymptoticEfficiencyExampleRatioMatching} recovers Besag's pseudolikelihood.
For this setting, the consistency and asymptotic normality of the corresponding estimator were established by \citet{Hyvarinen2006} and \citet{Nguyen2016}.
The proof is given in Appendix~\ref{sec:ProofExampleRatioMatching}.

\begin{proposition}
    \label{prp:AsymptoticEfficiencyExampleRatioMatching}
    Let $\L$ and $\L^\ast$ be defined in \eqref{eq:FlipOperator} and \eqref{eq:FlipOperatorAdjoint}, respectively,
    and let $h(x,y)\coloneqq\sum_{j=1}^{p}\{(1+y_j)\log(1+y_j)-y_j\log y_j\}/2$.
    Consider the Boltzmann machine
    \[
        p_{\theta}(x)\coloneqq \frac{1}{Z(\theta)}\exp\left(\sum_{j=1}^{p}x_j\eta_j+\sum_{j<k}x_jx_k\eta_{jk}\right)
    \]
    where $\theta=(\eta_1,\dots,\eta_p,\eta_{12},\dots,\eta_{p-1,p})^\top\in\Theta\coloneqq\R^d$ with $d=(p^2+p)/2$.
    If $p\leq3$, $\hat\theta$ is asymptotically efficient for any $\theta^\star\in\Theta$.
\end{proposition}

\subsection{Marginalization operator and Besag's pseudolikelihood}
\label{sec:ExamplesPseudoLikelihood}

Let $(\X_j,\mathscr{B}_j,\mathrm{d}x_j)$ for $j=1,\dots,p\, (p\geq2)$ be measure spaces, and consider the product space 
$(\X,\mathscr{B},\mathrm{d}x)\coloneqq(\prod_{j=1}^{p}\X_j,\bigotimes_{j=1}^{p}\mathscr{B}_j,\bigotimes_{j=1}^{p}\mathrm{d}x_j)$.
Define
\begin{equation}
    \label{eq:MarginalizationOperator}
    \L f(x)\coloneqq\left(\int_{\X_1}f(x)\d x_1,\dots,\int_{\X_p}f(x)\d x_p\right)^\top.
\end{equation}
By Lemma~4 of \citet{Lyu2009}, $\L^\ast$ is given by
\begin{equation}
    \label{eq:MarginalizationOperatorAdjoint}    
    \L^\ast V(x)\coloneqq\sum_{j=1}^{p}\int_{\X_j}V_j(x)\d x_j.
\end{equation}

In this setting, Besag's pseudolikelihood \citep{Besag1975} is recovered by choosing $h(x,y)\coloneqq\sum_{j=1}^{p}\log y_j$.
The corresponding scoring rule $S$, Stein operator $\A_{\theta}$, divergence $D(p_{\theta^\star}\,\|\,p_{\theta})$, and Dirichlet form $\EE_{\theta}$ are given by
\begin{align*}
    S(x,p_{\theta})&=
    -\sum_{j=1}^{p}\log p_{\theta}(x_j\mid x_{-j}),\\
    \A_{\theta}f(x)&=
    \sum_{j=1}^{p}\left(\E_{\theta}[f(X)\mid X_{-j}=x_{-j}]-f(x)\right),\\
    D(p_{\theta^\star}\,\|\,p_{\theta})&=\sum_{j=1}^{p}\E_{\theta^\star}\left[\log \frac{p_{\theta^\star}(X_j\mid X_{-j})}{p_{\theta}(X_j\mid X_{-j})}\right],\\
    \EE_{\theta}(f,g)&=
    \sum_{j=1}^{p}\E_{\theta}[\operatorname{Cov}_{\theta}(f(X),g(X)\mid X_{-j})],
\end{align*}
where $x_{-j}\coloneqq(x_1,\dots,x_{j-1},x_{j+1},\dots,x_{p})$ and $p_{\theta}(x_j\mid x_{-j})\coloneqq p_{\theta}(x)/\int_{\X_{j}}p_{\theta}(x)\d x_{j}$.

Note that when $\X=\{\pm1\}^{p}$, the marginalization operator is closely related to the flip operator.
Indeed, for each $j=1,\dots,p$,
$\L^\mathrm{marg}f(x)_j=\L^\mathrm{flip}f(x)_j+f(x)$, and therefore the two operators yield the same $\D_{\theta}$.
This relationship can also be understood through the choice of the neighborhood system; see Appendix~\ref{sec:ExamplesLocalScoringRuleDiscrete}.

\begin{remark}[Comparison with \citet{Fischer2025graph}]
    \citet{Fischer2025graph} developed Stein's method of moments for local dependency exponential random graph models using a Glauber dynamics Stein operator \citep{Reinert2019}.
    They showed that, by choosing the Fisher score functions as test functions, the resulting estimator coincides with the maximum pseudolikelihood estimator.
    From the perspective of our framework, this Stein operator is naturally induced by Besag's pseudolikelihood.
\end{remark}

The following proposition gives an alternative sufficient condition for $(\A_{\theta},\F_{\theta})$ to be a Stein pair.
The proof is given in Appendix~\ref{sec:ProofExamplePseudoLikelihood}.

\begin{proposition}
    \label{prp:SteinFormContractionPseudoLikelihood}
    Let $\L$ and $\L^\ast$ be defined in \eqref{eq:MarginalizationOperator} and \eqref{eq:MarginalizationOperatorAdjoint}, respectively,
    and let $h(x,y)\coloneqq\sum_{j=1}^{p}\log y_j$.
    Suppose that there exists a strictly increasing function $\Psi\colon(0,\infty)\to\R$ such that $\Psi(p_{\theta^\star}(\,\cdot\,)/p_{\theta}(\,\cdot\,))\in\F_{\theta}$ for all $\theta^\star,\theta\in\Theta$.
    Then $(\A_{\theta},\F_{\theta})$ is a Stein pair.
\end{proposition}

\subsection{Noise-weighted integral operator and noise contrastive estimation}
\label{sec:ExamplesNCE}
Let $n(x)>0$ be a probability density on $\X$, and $\L f(x)\coloneqq n(x)\int_\X f(x)\d x$.
$\L^\ast$ is given by $\L^\ast f(x)\coloneqq \E_n[f(W)]$, where $W\sim n(\,\cdot\,)$.
Define $h(x,y)\coloneqq (1+\nu y)\log (1+\nu y)-\nu y\log(\nu y)$, where $\nu>0$.
The corresponding scoring rule $S$, Stein operator $\A_{\theta}$, divergence $D(p_{\theta^\star}\,\|\,p_{\theta})$, and Dirichlet form $\EE_{\theta}$ are given by
\begin{align*}
    S(x,p_{\theta})&=-\log(1-w_{\theta}(x))-\nu\E_n[\log w_{\theta}(W)],\\
    \A_{\theta}f(x)
    &=-\E_{\theta}\left[w_{\theta}(X)\left(\E_{\theta}[f(X)]-f(X)\right)\right]+w_{\theta}(x)\left(\E_{\theta}[f(X)]-f(x)\right),\\
    D(p_{\theta^\star}\,\|\,p_{\theta})&=\E_{\theta^{\star}}\left[\log\frac{1-w_{\theta^{\star}}(X)}{1-w_{\theta}(X)}\right]+\nu\E_{n}\left[\log\frac{w_{\theta^{\star}}(W)}{w_{\theta}(W)}\right],\\
    \EE_{\theta}(f,g)&=\E_{\theta}\left[w_{\theta}(X)
    \left(f(X)-\E_{\theta}[f(X)]\right)\left(g(X)-\E_{\theta}[g(X)]\right)\right],
\end{align*}
where $w_{\theta}(x)\coloneqq \nu n(x)/(p_{\theta}(x)+\nu n(x))$.
Although this scoring rule depends on the normalizing constant $Z(\theta)$, introducing a free nuisance parameter $c$ for $-\log Z(\theta)$ and approximating the expectation with respect to $n(\,\cdot\,)$ by Monte Carlo recovers NCE \citep{Gutmann2010,Gutmann2012}.

\begin{remark}[Connection between the marginalization operator and noise contrastive estimation]
    \label{rem:NCEMarginalization}
    The NCE loss can also be derived using the marginalization operator defined in \eqref{eq:MarginalizationOperator}, rather than the noise-weighted integral operator.
    Let $\X\coloneqq\X'\times\{0,1\}$ and consider the parametric model defined by
    \[
    p_{\theta}(x,0)\coloneqq\frac{\nu}{1+\nu}n(x),\quad
    p_{\theta}(x,1)\coloneqq\frac{1}{1+\nu}q_{\theta}(x),
    \]
    where $q_{\theta}(x)>0$ is a probability density of interest on $\X'$, 
    $n(x)>0$ is an auxiliary probability density on $\X'$, and $\nu>0$.
    
    Let $\L$ be the marginalization operator defined in \eqref{eq:MarginalizationOperator} and define $h((x,z),y)\coloneqq\log y_2$.
    The corresponding scoring rule is given by
    \[
    S((x,z),p_{\theta})=-(1-z)\log w_{\theta}(x)-z\log(1-w_{\theta}(x)),
    \]
    where $w_{\theta}(x)\coloneqq \nu n(x)/(q_{\theta}(x)+\nu n(x))$.
    In this setting, the NCE loss is recovered by replacing $-\log Z(\theta)$ with a free nuisance parameter $c$.
    Thus, NCE can also be viewed as a binary conditional scoring rule induced by marginalization over the label variable.
\end{remark}

The following proposition gives an alternative sufficient condition for $(\A_{\theta},\F_{\theta})$ to be a Stein pair.
The proof is given in Appendix~\ref{sec:ProofExampleNCE}.

\begin{proposition}
    \label{prp:SteinFormContractionNCE}
    Let $\L f(x)\coloneqq n(x)\int_\X f(x)\d x$ and $\L^\ast f(x)\coloneqq\E_{n}[f(W)]$.
    Suppose $h_{yy}(x,y)<0$ and $1_{\{p_{\theta^\star}/p_{\theta}>1\}}(\,\cdot\,)\in\F_{\theta}$ for all $\theta^\star,\theta\in\Theta$.
    Then $(\A_{\theta},\F_{\theta})$ is a Stein pair.
\end{proposition}

\begin{remark}[Comparison with \citet{Fujii2026}]
    \citet{Fujii2026} generalized NCE using Bregman divergences between posterior class probabilities.
    Their risk functions are recovered by choosing
    \[
    h(x,y)\coloneqq-(1+\nu y)\left(\varphi\left(\frac{1}{1+\nu y}\right)+\varphi\left(\frac{\nu y}{1+\nu y}\right)\right),
    \]
    where $\varphi\colon(0,1)\to\R$ is a strictly convex function.
\end{remark}

The following proposition provides a closed-form expression for the Stein score functions corresponding to NCE.
Note that this result does not by itself establish the asymptotic efficiency of NCE since $-\log Z(\theta)$ is not treated as a nuisance parameter in this formulation.
The proof is given in Appendix~\ref{sec:ProofExampleNCE}.

\begin{proposition}
    \label{prp:SteinScoreNCE}
    For $j=1,\dots,d$, the Stein score functions corresponding to noise contrastive estimation are given by
    $\psi_{\theta,j}(x)=(\partial_{\theta_j}\log p_{\theta}(x)-C_{\theta,j})/{w_{\theta}(x)}$,
    where $w_{\theta}(x)\coloneqq \nu n(x)/(p_{\theta}(x)+\nu n(x))$ and 
    $C_{\theta,j}\coloneqq\E_{\theta}[{\partial_{\theta_j}\log p_{\theta}(X)}/{w_{\theta}(X)}]\E_{\theta}[{1}/{w_{\theta}(X)}]^{-1}$.
    Here, we assume that the quantities $C_{\theta,j}$ are well defined and that $\psi_{\theta,j}\in\operatorname{Dom}\A_{\theta}$ for $j=1,\dots,d$.
\end{proposition}

\begin{remark}[Comparison with \citet{Uehara2020}]
    For asymptotically efficient estimation without the normalizing constant,
    \citet{Uehara2020} proposed self density-ratio matching estimators that replace the unknown true density $p_{\theta^\star}$ by a nonparametric density estimator.
    In our framework, their construction is closely related to the ideal choice $n(x)=p_{\theta^\star}(x)$.
    Indeed, under this choice, the Stein score functions are given by
    $\psi_{\theta^\star,j}=(1+1/{\nu})\partial_{\theta_j}\log p_{\theta^\star}$,
    so that part (2) of Theorem~\ref{thm:AsymptoticEfficiency} is satisfied.
\end{remark}

\subsection{Integral operator and MLE}
\label{sec:ExamplesMLE}
Let $\L f(x)\coloneqq \int_{\X}f(x)\d x$.
$\L^\ast$ is given by $\L^\ast\coloneqq\L$.
In this setting, MLE (log score) is recovered by choosing $h(x,y)\coloneqq\log y$.
The corresponding scoring rule $S$, Stein operator $\A_{\theta}$, divergence $D(p_{\theta^\star}\,\|\,p_{\theta})$, and Dirichlet form $\EE_{\theta}$ are given by
\begin{align*}
    S(x,p_{\theta})&=
    -\log p_{\theta}(x),\\
    \A_{\theta}f(x)&=\E_{\theta}[f(X)]-f(x),\\
    D(p_{\theta^\star}\,\|\,p_{\theta})&=\E_{\theta^\star}[\log p_{\theta^\star}(X)-\log p_{\theta}(X)],\\
    \EE_{\theta}(f,g)&=\operatorname{Cov}_{\theta}(f(X),g(X)).
\end{align*}
Note that this scoring rule depends on the normalizing constant, since $\L p_{\theta}/p_{\theta}=1/p_{\theta}$ depends on it.
In this setting, the Stein score functions coincide with the Fisher score functions.

The following proposition gives an alternative sufficient condition for $(\A_{\theta},\F_{\theta})$ to be a Stein pair.
The proof is given in Appendix~\ref{sec:ProofExampleMLE}.

\begin{proposition}
    \label{prp:SteinFormContractionMLE}
    Let $\L f(x)\coloneqq\int_\X f(x)\d x$ and $\L^\ast\coloneqq\L$.
    Suppose $h_{yy}(x,y)<0$ and $1_{\{p_{\theta^\star}/p_{\theta}>1\}}(\,\cdot\,)\in\F_{\theta}$ for all $\theta^\star,\theta\in\Theta$.
    Then $(\A_{\theta},\F_{\theta})$ is a Stein pair.
\end{proposition}

\section{Conclusion}
We have developed a unified framework that connects proper scoring rules and Stein's method.
Our class of proper scoring rules encompasses the principal methods for estimation in unnormalized models, including score matching, ratio matching, pseudolikelihood, and noise contrastive estimation.
We have shown that these scoring rules naturally induce Stein operators through their estimating equations, which provides a new principle for constructing Stein operators.
Using the geometry induced by these operators, we have established a unified characterization of the asymptotic efficiency of the resulting estimators.

\section*{Acknowledgements}
The authors would like to thank Rikiya Okazaki for helpful comments.
This work was supported by JSPS KAKENHI Grant Numbers
JP24K15120, JP24H00247, JP26K02871, JP26K23255, JP26K00325,
JP26K02869, by JST CREST Grant Number JPMJCR26E3,
by MEXT STAR-E NEXT Project (Japan Grant Number JPJ013735).

\bibliography{cite.bib}

@article{Ackley1985,
author = {Ackley, David H. and Hinton, Geoffrey E. and Sejnowski, Terrence J.},
title = {A learning algorithm for {{Boltzmann}} machines},
journal = {Cognitive Science},
volume = {9},
number = {1},
pages = {147-169},
doi = {https://doi.org/10.1207/s15516709cog0901\_7},
url = {https://onlinelibrary.wiley.com/doi/abs/10.1207/s15516709cog0901_7},
year = {1985}
}

@article{Amari2024,
  author  = {Amari, {{Shun-ichi}} and Matsuda, Takeru},
  title   = {Information geometry of {{Wasserstein}} statistics on shapes and affine deformations},
  journal = {Information Geometry},
  volume  = {7},
  pages   = {285--309},
  year    = {2024},
}

@article{Anastasiou2023,
author = {Anastasiou, Andreas and Barp, Alessandro and Briol, Fran{\c{c}}ois-Xavier and Ebner, Bruno and Gaunt, Robert E. and Ghaderinezhad, Fatemeh and Gorham, Jackson and Gretton, Arthur and Ley, Christophe and Liu, Qiang and Mackey, Lester and Oates, Chris J. and Reinert, Gesine and Swan, Yvik},
title = {{{Stein's}} method meets computational statistics: {{A}} review of some recent developments},
volume = {38},
journal = {Statistical Science},
number = {1},
publisher = {Institute of Mathematical Statistics},
pages = {120--139},
year = {2023},
doi = {10.1214/22-STS863},
URL = {https://doi.org/10.1214/22-STS863}
}

@article{Barp2019,
  title = {Minimum {{Stein discrepancy estimators}}},
  journal = {Advances in {{Neural Information Processing Systems}}},
  author = {Barp, Alessandro and Briol, Francois-Xavier and Duncan, Andrew and Girolami, Mark and Mackey, Lester},
  year = {2019},
  volume = {32},
  publisher = {Curran Associates, Inc.},
  urldate = {2025-06-15},
}

@article{Besag1974,
 author = {Besag, Julian},
 journal = {Journal of the Royal Statistical Society. Series B (Methodological)},
 number = {2},
 pages = {192--225},
 publisher = {[Royal Statistical Society, Oxford University Press]},
 title = {Spatial interaction and the statistical analysis of lattice systems},
 urldate = {2026-08-31},
 volume = {36},
 year = {1974}
}

@article{Besag1975,
 author = {Besag, Julian},
 journal = {Journal of the Royal Statistical Society. Series D (The Statistician)},
 number = {3},
 pages = {179--195},
 publisher = {[Royal Statistical Society, Wiley]},
 title = {Statistical Analysis of Non-Lattice Data},
 urldate = {2026-09-03},
 volume = {24},
 year = {1975}
}

@article{Chafai2004,
author = {Djalil Chafa{\"i}},
title= {Entropies, convexity, and functional inequalities: {{On}} {{$\Phi$}}-entropies and {{$\Phi$-Sobolev}} inequalities},
volume = {44},
journal = {Journal of Mathematics of Kyoto University},
number = {2},
publisher = {Duke University Press},
pages = {325--363},
year = {2004},
doi = {10.1215/kjm/1250283556},
URL = {https://doi.org/10.1215/kjm/1250283556}
}

@book{Chikuse2003,
  author    = {Chikuse, Yasuko},
  title     = {Statistics on Special Manifolds},
  series    = {Lecture Notes in Statistics},
  volume    = {174},
  publisher = {Springer},
  address   = {New York},
  year      = {2003},
  doi       = {10.1007/978-0-387-21540-2}
}

@inproceedings{Dawid2005,
  title = {The geometry of decision theory},
  booktitle = {The {{2nd International Symposium}} on {{Information Geometry}}  and  its {{Applications}}},
  author = {Dawid, A. Philip and Lauritzen, Steffen},
  year = 2005,
  pages = {22--28},
  langid = {english},
}

@article{Dawid2007,
  author  = {Dawid, A. Philip},
  title   = {The geometry of proper scoring rules},
  journal = {Annals of the Institute of Statistical Mathematics},
  volume  = {59},
  number  = {1},
  pages   = {77--93},
  year    = {2007},
  doi     = {10.1007/s10463-006-0099-8}
}

@article{Dawid2012,
author = {Dawid, A. Philip and Lauritzen, Steffen and Parry, Matthew},
title = {Proper local scoring rules on discrete sample spaces},
volume = {40},
journal = {The Annals of Statistics},
number = {1},
publisher = {Institute of Mathematical Statistics},
pages = {593--608},
year = {2012},
doi = {10.1214/12-AOS972},
URL = {https://doi.org/10.1214/12-AOS972}
}

@article{Dawid2016,
author = {Dawid, A. Philip and Musio, Monica and Ventura, Laura},
title = {Minimum scoring rule inference},
journal = {Scandinavian Journal of Statistics},
volume = {43},
number = {1},
pages = {123--138},
year = {2016}
}

@article{Ebner2025,
  title = {Stein's method of moments},
  author = {Ebner, Bruno and Fischer, Adrian and Gaunt, Robert E. and Picker, Babette and Swan, Yvik},
  year = {2025},
  volume = {52},
  number = {4},
  pages = {1594--1624},
  journal = {Scandinavian Journal of Statistics},
  publisher = {Wiley},
  issn = {0303-6898, 1467-9469},
  doi = {10.1111/sjos.70003},
}

@article{Eguchi1992,
author = {Shinto Eguchi},
title = {{Geometry of minimum contrast}},
volume = {22},
journal = {Hiroshima Mathematical Journal},
number = {3},
publisher = {Hiroshima University, Mathematics Program},
pages = {631--647},
year = {1992},
doi = {10.32917/hmj/1206128508},
URL = {https://doi.org/10.32917/hmj/1206128508}
}

@article{Eguchi2026,
    author = {Shinto Eguchi},
    title = {Robust inference using density-powered {{Stein}} operators},
    journal = {Information Geometry},
    year = {2026},
}

@article{Ehm2012,
author = {Werner Ehm and Tilmann Gneiting},
title = {Local proper scoring rules of order two},
volume = {40},
journal = {The Annals of Statistics},
number = {1},
publisher = {Institute of Mathematical Statistics},
pages = {609--637},
year = {2012},
doi = {10.1214/12-AOS973},
URL = {https://doi.org/10.1214/12-AOS973}
}

@article{Fischer2025discrete,
  title={New closed-form estimators for discrete distributions},
  author={Fischer, Adrian},
  journal={arXiv preprint arXiv:2510.18503},
  year={2025}
}

@article{Fischer2025truncated,
  title = {{{Stein's}} method of moments for truncated multivariate distributions},
  author = {Fischer, Adrian and Gaunt, Robert E. and Swan, Yvik},
  year = {2025},
  journal = {Electronic Journal of Statistics},
  volume = {19},
  number = {1},
  pages = {1784--1808},
  publisher = {Institute of Mathematical Statistics},
  doi = {10.1214/25-ejs2373},
}

@article{Fischer2025graph,
  title={{{Stein's}} method of moment estimators for local dependency exponential random graph models},
  author={Fischer, Adrian and Reinert, Gesine and Xu, Wenkai},
  journal={arXiv preprint arXiv:2503.13191},
  year={2025}
}

@article{Fischer2026,
author = {Fischer, Adrian and Gaunt, Robert E. and Swan, Yvik},
title = {{{Stein's}} method of moments on the sphere},
volume = {32},
journal = {Bernoulli},
number = {2},
publisher = {Bernoulli Society for Mathematical Statistics and Probability},
pages = {1186--1212},
year = {2026},
doi = {10.3150/25-BEJ1898},
URL = {https://doi.org/10.3150/25-BEJ1898}
}

@article{Fukushi2026,
  author  = {Fukushi, Ayumu and Nakanishi-Ohno, Yoshinori and Matsuda, Takeru},
  title   = {Flatness of location-scale-shape models under the {{Wasserstein}} metric},
  journal = {Information Geometry},
  year    = {2026},
}

@InProceedings{Fujii2026,
  title = {{Bregman} contrastive estimation as a general framework for learning unnormalized models: {{Efficiency}}, robustness and optimal noise},
  author =  {Fujii, Yuto and Sasaki, Hiroaki and Kanamori, Takafumi},
  booktitle = {Proceedings of the 42nd Conference on Uncertainty in Artificial Intelligence (UAI 2026)},
  pages = {1604--1627},
  year = 	 {2026}
}

@article{Fujisawa2008,
title = {Robust parameter estimation with a small bias against heavy contamination},
journal = {Journal of Multivariate Analysis},
volume = {99},
number = {9},
pages = {2053--2081},
year = {2008},
author = {Hironori Fujisawa and Shinto Eguchi},
}

@article{Gneiting2007,
author = {Tilmann Gneiting and Adrian E. Raftery},
title = {Strictly proper scoring rules, prediction, and estimation},
journal = {Journal of the American Statistical Association},
volume = {102},
number = {477},
pages = {359--378},
year = {2007},
publisher = {Taylor \& Francis},
}

@article{Gorham2015,
 title = {Measuring sample quality with {{Stein's}} method},
 author = {Gorham, Jackson and Mackey, Lester},
 journal = {Advances in Neural Information Processing Systems},
 publisher = {Curran Associates, Inc.},
 volume = {28},
 year = {2015}
}

@article{Gorham2019,
 author = {Gorham, Jackson and Duncan, Andrew B. and Vollmer, Sebastian J. and Mackey, Lester},
 journal = {The Annals of Applied Probability},
 number = {5},
 pages = {2884--2928},
 publisher = {Institute of Mathematical Statistics},
 title = {Measuring sample quality with diffusions},
 volume = {29},
 year = {2019}
}

@InProceedings{Grathwohl2020,
  title = {Learning the {{Stein}} discrepancy for training and evaluating energy-based models without Sampling},
  author = {Grathwohl, Will and Wang, Kuan-Chieh and Jacobsen, Joern-Henrik and Duvenaud, David and Zemel, Richard},
  booktitle = {Proceedings of the 37th International Conference on Machine Learning (ICML 2020)},
  pages = {3732--3747},
  year = {2020},
}

@article{Glynn1996,
 author = {Peter W. Glynn and Sean P. Meyn},
 journal = {The Annals of Probability},
 number = {2},
 pages = {916--931},
 publisher = {Institute of Mathematical Statistics},
 title = {A {{Liapounov}} bound for solutions of the {{Poisson}} equation},
 volume = {24},
 year = {1996}
}

@inproceedings{Gutmann2010,
  title = 	 {Noise-contrastive estimation: {{A}} new estimation principle for unnormalized statistical models},
  author = 	 {Gutmann, Michael U. and Hyv{{\"a}}rinen, Aapo},
  booktitle = {Proceedings of the 13th International Conference on Artificial Intelligence and Statistics (AISTATS 2010)},
  pages = 	 {297--304},
  year = 	 {2010}
}

@inproceedings{Gutmann2011,
author = {Gutmann, Michael U. and Hirayama, {{Jun-ichiro}}},
title = {Bregman divergence as general framework to estimate unnormalized statistical models},
year = {2011},
isbn = {9780974903972},
booktitle = {Proceedings of the 27th Conference on Uncertainty in Artificial Intelligence (UAI 2011)},
pages = {283--290},
numpages = {8},
}

@article{Gutmann2012,
  author  = {Michael U. Gutmann and Aapo Hyv{\"a}rinen},
  title   = {Noise-contrastive estimation of unnormalized statistical Models, with applications to natural image statistics},
  journal = {Journal of Machine Learning Research},
  year    = {2012},
  volume  = {13},
  pages   = {307--361},
  url     = {http://jmlr.org/papers/v13/gutmann12a.html}
}

@article{Hyvarinen2005,
  title = {Estimation of {{non-normalized statistical models}} by {{score matching}}},
  author = {Hyv{\"a}rinen, Aapo},
  editor = {Dayan, Peter},
  year = {2005},
  journal = {Journal of Machine Learning Research},
  volume = {6},
  pages = {695--709},
}

@article{Hyvarinen2006,
    author = {Hyv{\"a}rinen, Aapo},
    title = {Consistency of pseudolikelihood estimation of fully visible {{Boltzmann}} machines},
    journal = {Neural Computation},
    volume = {18},
    number = {10},
    pages = {2283--2292},
    year = {2006},
    month = {10},
    issn = {0899-7667},
    doi = {10.1162/neco.2006.18.10.2283},
    url = {https://doi.org/10.1162/neco.2006.18.10.2283},
    eprint = {https://direct.mit.edu/neco/article-pdf/18/10/2283/816603/neco.2006.18.10.2283.pdf},
}

@article{Hyvarinen2007,
  title = {Some extensions of score matching},
  author = {Hyv{\"a}rinen, Aapo},
  year = {2007},
  journal = {Computational Statistics \& Data Analysis},
  volume = {51},
  number = {5},
  pages = {2499--2512},
  issn = {01679473},
  doi = {10.1016/j.csda.2006.09.003},
}

@article{Kanamori2015,
  title = {Robust estimation under heavy contamination using unnormalized models},
  author = {Kanamori, Takafumi and Fujisawa, Hironori},
  year = 2015,
  month = sep,
  journal = {Biometrika},
  volume = {102},
  number = {3},
  pages = {559--572},
  issn = {0006-3444, 1464-3510},
  doi = {10.1093/biomet/asv014},
  urldate = {2025-06-04},
  langid = {english},
}

@article{Kanamori2017,
title = {Graph-based composite local {{Bregman}} divergences on discrete sample spaces},
journal = {Neural Networks},
volume = {95},
pages = {44--56},
year = {2017},
issn = {0893-6080},
author = {Takafumi Kanamori and Takashi Takenouchi},
}

@inproceedings{Koehler2023,
  author={Frederic Koehler and Alexander Heckett and Andrej Risteski},
  title={Statistical efficiency of score matching: {{The}} view from isoperimetry},
  booktitle={The 11th International Conference on Learning Representations (ICLR 2023) },
  year={2023},
  url={https://openreview.net/forum?id=TD7AnQjNzR6}
}

@article{Kume2026,
  title={On {{Stein's}} method of moments and generalized score matching},
  author={Kume, Alfred and Walker, Stephen G.},
  journal={arXiv preprint arXiv:2602.06482},
  year={2026}
}

@article{Lai2025,
  title = {The principles of diffusion models},
  author = {Lai, Chieh-Hsin and Song, Yang and Kim, Dongjun and Mitsufuji, Yuki and Ermon, Stefano},
  year = {2025},
  journal={arXiv preprint arXiv:2510.21890},
}

@article{Li2023,
  title={Wasserstein information matrix},
  author={Li, Wuchen and Zhao, Jiaxi},
  journal={Information Geometry},
  volume={6},
  number={1},
  pages={203--255},
  year={2023},
  publisher={Springer}
}

@article{Liu2022,
  title = {Estimating {{density models}} with {{truncation boundaries}} using {{score matching}}},
  author = {Liu, Song and Kanamori, Takafumi and Williams, Daniel J.},
  year = {2022},
  volume = {23},
  journal = {Journal of Machine Learning Research},
  pages = {1--38},
}

@article{Liu2026,
  title={Probabilistic inference and learning with {{Stein's}} method},
  author={Liu, Qiang and Mackey, Lester and Oates, Chris},
  journal={arXiv preprint arXiv:2603.07467},
  year={2026}
}

@incollection{LeCun2006,
  author    = {LeCun, Yann and Chopra, Sumit and Hadsell, Raia and Ranzato, Marc'Aurelio and Huang, Fu J.},
  title     = {A Tutorial on Energy-Based Learning},
  booktitle = {Predicting Structured Data},
  publisher = {MIT Press},
  address   = {Cambridge, MA},
  year      = {2006},
  chapter   = {10},
  pages     = {191--246}
}

@article{Ley2013,
author = {Ley, Christophe and Swan, Yvik},
title = {{Stein's density approach and information inequalities}},
volume = {18},
journal = {Electronic Communications in Probability},
publisher = {Institute of Mathematical Statistics and Bernoulli Society},
pages = {1--14},
year = {2013},
doi = {10.1214/ECP.v18-2578},
URL = {https://doi.org/10.1214/ECP.v18-2578}
}

@article{Ley2016bjps,
 title = {Parametric {{Stein}} operators and variance bounds},
 author = {Ley, Christophe and Swan, Yvik},
 journal = {Brazilian Journal of Probability and Statistics},
 number = {2},
 pages = {171--195},
 publisher = {[Brazilian Statistical Association, Institute of Mathematical Statistics]},
 volume = {30},
 year = {2016}
}

@article{Ley2016spl,
title = {A general parametric {{Stein}} characterization},
author = {Ley, Christophe and Swan, Yvik},
journal = {Statistics \& Probability Letters},
volume = {111},
pages = {67--71},
year = {2016},
issn = {0167-7152},
doi = {https://doi.org/10.1016/j.spl.2016.01.008},
url = {https://www.sciencedirect.com/science/article/pii/S0167715216000079},
}

@inproceedings{Lyu2009,
author = {Lyu, Siwei},
title = {Interpretation and generalization of score matching},
year = {2009},
isbn = {9780974903958},
booktitle = {Proceedings of the 25th Uncertainty in Artificial Intelligence Conference (UAI 2009)},
pages = {359--366},
numpages = {8},
}

@book{Mardia1999,
  title={{Directional Statistics}},
  author={Mardia, Kanti V. and Jupp, Peter E.},
  year={1999},
  publisher={John Wiley \& Sons}
}

@article{Mardia2016,
  title = {Score matching estimators for directional distributions},
  author = {Mardia, Kanti V. and Kent, John T. and Laha, Arnab K.},
  year = {2016},
  journal={arXiv preprint arXiv:1604.08470},
}

@inproceedings{Marlin2011,
author = {Marlin, Benjamin M. and de Freitas, Nando},
title = {Asymptotic efficiency of deterministic estimators for discrete energy-based models: ratio matching and pseudolikelihood},
year = {2011},
booktitle = {Proceedings of the 27th Conference on Uncertainty in Artificial Intelligence (UAI 2011)},
pages = {497--505},
}

@inproceedings{Matsuda2019,
  title={Estimation of non-normalized mixture models},
  author={Matsuda, Takeru and Hyv{\"a}rinen, Aapo},
  booktitle={Proceedings of The 22nd International Conference on Artificial Intelligence and Statistics (AISTATS 2019)},
  pages={2555--2563},
  year={2019},
}

@article{Meng2022,
  title={Concrete score matching: {{Generalized}} score matching for discrete data},
  author={Meng, Chenlin and Choi, Kristy and Song, Jiaming and Ermon, Stefano},
  journal={Advances in Neural Information Processing Systems},
  volume={35},
  pages={34532--34545},
  year={2022}
}

@article{Mijoule2023,
  title = {Stein's density method for multivariate continuous distributions},
  author = {Mijoule, Guillaume and Rai{\v c}, Martin and Reinert, Gesine and Swan, Yvik},
  year = {2023},
  journal = {Electronic Journal of Probability},
  volume = {28},
  number = {59},
  pages = {1--40},
  publisher = {Institute of Mathematical Statistics},
  issn = {1083-6489},
  doi = {10.1214/22-ejp883},
}

@article{Nagai2026,
  title={The geometry of {{Stein's}} method of moments: {{A}} canonical decomposition via score matching},
  author={Nagai, Mitsuki and Yano, Keisuke},
  journal={arXiv preprint arXiv:2603.12843},
  year={2026}
}

@article{Nishimori2025,
  title={{On the attainment of the Wasserstein--Cramer--Rao lower bound}},
  author={Nishimori, Hayato and Matsuda, Takeru},
  journal={Information Geometry},
  year={2025},
  publisher={Springer}
}

@article{Nguyen2016,
  author={Nguyen, Hien D. and Wood, Ian A.},
  journal={IEEE Transactions on Neural Networks and Learning Systems}, 
  title={Asymptotic normality of the maximum pseudolikelihood estimator for fully visible {{Boltzmann}} machines}, 
  year={2016},
  volume={27},
  number={4},
  pages={897--902},
  doi={10.1109/TNNLS.2015.2425898}
}

@book{ODonnell2014, 
  author={O'Donnell, Ryan}, 
  title={Analysis of Boolean Functions}, 
  publisher={Cambridge University Press}, 
  place={Cambridge}, 
  year={2014}
}

@article{Parry2012,
author = {Parry, Matthew and Dawid, A. Philip and Lauritzen, Steffen},
title = {Proper local scoring rules},
volume = {40},
journal = {The Annals of Statistics},
number = {1},
publisher = {Institute of Mathematical Statistics},
pages = {561--592},
year = {2012},
doi = {10.1214/12-AOS971},
URL = {https://doi.org/10.1214/12-AOS971}
}

@InProceedings{Qin2024,
  title = {Fit like you sample: {{Sample}}-efficient generalized score matching from fast mixing diffusions},
  author = {Qin, Yilong and Risteski, Andrej},
  booktitle = {Proceedings of 37th Conference on Learning Theory (COLT 2024)},
  pages = {4413--4457},
  year = {2024},
}

@article{Scealy2023,
  title = {Score {{matching}} for {{compositional distributions}}},
  author = {Scealy, Janice L. and Wood, Andrew T. A.},
  year = {2023},
  journal = {Journal of the American Statistical Association},
  volume = {118},
  number = {543},
  pages = {1811--1823},
  issn = {0162-1459, 1537-274X},
  doi = {10.1080/01621459.2021.2016422},
  urldate = {2025-05-10},
}

@article{Song2021,
  title={How to train your energy-based models},
  author={Song, Yang and Kingma, Diederik P.},
  journal={arXiv preprint arXiv:2101.03288},
  year={2021}
}

@article{Sonobe2026,
  title={Contrastive {B}ayesian Inference for Unnormalized Models},
  author={Sonobe, Naruki and Sugasawa, Shonosuke and Mochihashi, Daichi and Matsuda, Takeru},
  journal={arXiv preprint arXiv:2603.09306},
  year={2026}
}

@inproceedings{Stein1972,
  title={A bound for the error in the normal approximation to the distribution of a sum of dependent random variables},
  author={Stein, Charles},
  booktitle={Proceedings of the 6th Berkeley symposium on mathematical statistics and probability, volume 2: Probability theory},
  pages={583--602},
  year={1972},
  organization={University of California Press}
}

@article{Reinert2019,
author={Reinert, Gesine and Ross, Nathan},
title={Approximating stationary distributions of fast mixing {G}lauber dynamics, with applications to exponential random graphs},
journal={The Annals of Applied Probability},
year={2019},
pages={3201--3229},
volume={29},
number={5}
}

@book{Rockafellar,
 author = {R. Tyrrell Rockafellar},
 publisher = {Princeton University Press},
 title = {Convex Analysis},
 year = {1970}
}

@article{Takasu2018,
  title = {Scoring rules for statistical models on spheres},
  author = {Takasu, Yuya and Yano, Keisuke and Komaki, Fumiyasu},
  year = {2018},
  journal = {Statistics \& Probability Letters},
  volume = {138},
  pages = {111--115},
  issn = {01677152},
  doi = {10.1016/j.spl.2018.02.054},
}

@InProceedings{Uehara2020,
  title = {A unified statistically efficient estimation framework for unnormalized models },
  author = {Uehara, Masatoshi and Kanamori, Takafumi and Takenouchi, Takashi and Matsuda, Takeru},
  booktitle = {Proceedings of the 23rd International Conference on Artificial Intelligence and Statistics (AISTATS 2020)},
  pages = {809--819},
  year = {2020},
}

@book{VanderVaart1998,
  title = {Asymptotic {{Statistics}}},
  author = {{Van der Vaart}, Aad W.},
  year = {1998},
  edition = {1},
  publisher = {Cambridge University Press},
  doi = {10.1017/CBO9780511802256},
  urldate = {2025-04-14},
  isbn = {978-0-511-80225-6 978-0-521-49603-2 978-0-521-78450-4},
}

@article{Waghmare2026,
  author = {Waghmare, Kartik and Ziegel, Johanna},
  journal = {Annual Review of Statistics and Its Application},
  title = {Proper scoring rules for estimation and forecast evaluation},
  volume = {13},
  pages = {271--296},
  year = {2026}
}

@inproceedings{Williams2022,
  title = {Score {{matching}} for {{truncated density estimation}} on a {{manifold}}},
  booktitle = {Proceedings of {{Topological}}, {{Algebraic}}, and {{Geometric Learning Workshops}} 2022 ({{TAG-ML}} 2022)},
  author = {Williams, Daniel J. and Liu, Song},
  year = {2022},
}

@InProceedings{Yang2018,
  title = {Goodness-of-fit testing for discrete distributions via {{Stein}} discrepancy},
  author = {Yang, Jiasen and Liu, Qiang and Rao, Vinayak and Neville, Jennifer},
  booktitle = {Proceedings of the 35th International Conference on Machine Learning (ICML 2018)},
  pages = {5561--5570},
  year = 	{2018},
}

@article{Yu2019,
  title = {Generalized {{score matching}} for {{non-negative data}}},
  author = {Yu, Shiqing and Drton, Mathias and Shojaie, Ali},
  year = {2019},
  journal = {Journal of Machine Learning Research},
  volume = {20},
  pages = {1--70},
}

\clearpage
\appendix
\section{Regularity conditions}
\label{sec:RegularityConditions}

For the scoring rule $S$ defined in \eqref{eq:scoringrule},
we impose the following regularity conditions on $\L$ and $h$ to ensure that all relevant expectations are finite and that the functions involved lie in the appropriate operator domains.

\begin{enumerate}
    \item $h(\,\cdot\,,\L p_{\theta}(\,\cdot\,)/p_{\theta}(\,\cdot\,))\in L^1(\P)$ for all $\theta\in\Theta$;
    \item $h_y(\,\cdot\,,\L p_{\theta}(\,\cdot\,)/p_{\theta}(\,\cdot\,))\in \operatorname{Dom}\L^\ast$ for all $\theta\in\Theta$;
    \item $\L^\ast[h_y(\,\cdot\,,\L p_{\theta}(\,\cdot\,)/p_{\theta}(\,\cdot\,))]\in L^1(\P)$ for all $\theta\in\Theta$;
    \item $\ang{h_y(\,\cdot\,,\L p_{\theta}(\,\cdot\,)/p_{\theta}(\,\cdot\,)),\L p_{\theta}(\,\cdot\,)/p_{\theta}(\,\cdot\,)}\in L^1(\P)$ for all $\theta\in\Theta$.
    \item $\ang{h_y(\,\cdot\,,\L p_{\theta}(\,\cdot\,)/p_{\theta}(\,\cdot\,)),\L p_{\theta^\star}(\,\cdot\,)/p_{\theta^\star}(\,\cdot\,)}\in L^1(p_{\theta^\star})$ for all $\theta,\theta^\star\in\Theta$.
\end{enumerate}

$\operatorname{Dom}\A_{\theta}\subset L^2(p_{\theta})$ is the space of functions $f$ satisfying

\begin{enumerate}
    \item $fp_{\theta}\in\operatorname{Dom}\L$;
    \item $h_{yy}\left(\,\cdot\,,{\L p_{\theta}(\,\cdot\,)}/{p_{\theta}(\,\cdot\,)}\right)\D_{\theta}f \in \operatorname{Dom}\L^\ast$;
    \item $\L^\ast\left[h_{yy}\left(\,\cdot\,,{\L p_{\theta}(\,\cdot\,)}/{p_{\theta}(\,\cdot\,)}\right)\D_{\theta}f\right] \in L^2(p_{\theta})$;
    \item $\ang*{h_{yy}\left(\,\cdot\,,{\L p_{\theta}(\,\cdot\,)}/{p_{\theta}(\,\cdot\,)}\right)\D_{\theta}f(\,\cdot\,),{\L p_{\theta}(\,\cdot\,)}/{p_{\theta}(\,\cdot\,)}}\in L^2(p_{\theta})$;
    \item $\ang*{h_{yy}\left(\,\cdot\,,{\L p_{\theta}(\,\cdot\,)}/{p_{\theta}(\,\cdot\,)}\right)\D_{\theta}f(\,\cdot\,),\D_{\theta}f(\,\cdot\,)}\in L^1(p_{\theta}).$
\end{enumerate}

Let $\F_{\theta}\subset\operatorname{Dom}\A_{\theta}$ be a subset such that, for every $f\in\F_{\theta}$,
\[
\L^\ast\left[h_{yy}\left(\,\cdot\,,\frac{\L p_{\theta}(\,\cdot\,)}{p_{\theta}(\,\cdot\,)}\right)\D_{\theta}f\right]\in L^{1}(\P)
\]
and 
\[
\ang*{h_{yy}\left(\,\cdot\,,\frac{\L p_{\theta}(\,\cdot\,)}{p_{\theta}(\,\cdot\,)}\right)\D_{\theta}f(\,\cdot\,),\frac{\L p_{\theta}(\,\cdot\,)}{p_{\theta}(\,\cdot\,)}}\in L^1(\P),
\]
which ensure that $\A_{\theta}f\in L^1(\P)$ for all $f\in\F_{\theta}$.

To provide a unified characterization of the asymptotic efficiency of the estimator $\hat\theta$,
we assume the following regularity conditions.

\begin{itemize}
    \item[R1.] The matrix $G(\theta)\in\R^{d\times d}$, with $(j,k)$-th entry
    \[
        G(\theta)_{jk}\coloneqq\E_{\theta}\left[\partial_{\theta_k}(\A_{\theta}\partial_{\theta_j}\log p_{\theta}(X))\right],
    \]
    is nonsingular at $\theta=\theta^\star$.
    \item[R2.] The Fisher information matrix $I(\theta)\in\R^{d\times d}$, with $(j,k)$-th entry
    \[
        I(\theta)_{jk}\coloneqq\E_{\theta}[\partial_{\theta_j}\log p_{\theta}(X)\partial_{\theta_k}\log p_{\theta}(X)],
    \]
    is nonsingular at $\theta=\theta^\star$.
    \item[R3.] Differentiation under the integral sign is valid at $\theta=\theta^\star$:
    \[
      \int_\X \partial_{\theta_k}(p_{\theta}(x)\A_{\theta}\partial_{\theta_j}\log p_{\theta}(x))\d x
        =\partial_{\theta_k}\E_{\theta}[\A_{\theta}\partial_{\theta_j}\log p_{\theta}(X)]
    \]
    \item[R4.] The estimator $\hat\theta$ defined in \eqref{eq:scoringruleestimator} with the scoring rule $S$ defined in \eqref{eq:scoringrule} is consistent and asymptotically linear; that is,
    \[
        \hat\theta-\theta^\star=-G(\theta^\star)^{-1}\frac{1}{n}\sum_{i=1}^{n}\begin{pmatrix}\A_{\theta^\star}\partial_{\theta_1}\log p_{\theta^\star}(X_i)\\\vdots\\\A_{\theta^\star}\partial_{\theta_d}\log p_{\theta^\star}(X_i)\end{pmatrix}+\smallop[n^{-1/2}].
    \]
    \item[R5.] The maximum likelihood estimator $\hat\theta_\mathrm{MLE}$ is consistent and asymptotically linear; that is,
    \[
        \hat\theta_\mathrm{MLE}-\theta^\star=I(\theta^\star)^{-1}\frac{1}{n}\sum_{i=1}^{n}\nabla_{\theta}\log p_{\theta^\star}(X_i)+\smallop[n^{-1/2}].
    \]
\end{itemize}

\section{Auxiliary results}
\label{sec:AuxiliaryResults}

In this section, we provide auxiliary results omitted from the main text.

\subsection{Sufficient conditions for Assumptions~\ref{assumption:L-identifiability} and \ref{assumption:SteinFormKernel}}
\label{sec:SuffCondForAsm}

We first introduce the following kernel condition on $\D_{\theta}$.
\begin{assumption}
    \label{assumption:DKernel}
    For all $\theta\in\Theta$ and all $f\in\Meas(\X;\R)$ such that $fp_{\theta}\in\operatorname{Dom}\L$, $\D_{\theta}f=0$ if and only if $f$ is constant.
\end{assumption}

The following proposition provides sufficient conditions for Assumptions~\ref{assumption:L-identifiability} and \ref{assumption:SteinFormKernel}.

\begin{proposition}
    \label{proposition:SuffCondForAsm}
    Suppose Assumption~\ref{assumption:DKernel} holds.
    Then Assumption~\ref{assumption:L-identifiability} is satisfied.
    Suppose further that $h_{yy}(x,y)$ is negative definite.
    Then Assumption~\ref{assumption:SteinFormKernel} is also satisfied.
\end{proposition}

\begin{proof}
    We first show that Assumption~\ref{assumption:L-identifiability} is satisfied.
    Fix an arbitrary $\theta^\star\in\Theta$.
    Observe that
    \[
        \D_{\theta}\left[\frac{p_{\theta^{\star}}}{p_{\theta}}\right]
        =\frac{p_{\theta^{\star}}}{p_{\theta}}\left(\frac{\L p_{\theta^\star}}{p_{\theta^\star}}-\frac{\L p_{\theta}}{p_{\theta}}\right).
    \]
    Since $p_{\theta^\star}\in\operatorname{Dom}\L$,
    Assumption~\ref{assumption:DKernel} applies to $p_{\theta^\star}/p_{\theta}$.
    Hence, $\L p_{\theta}/p_{\theta}=\L p_{\theta^\star}/p_{\theta^\star}$ holds if and only if $p_{\theta^\star}/p_{\theta}$ is constant, which implies $\theta=\theta^\star$.

    Next, we show that Assumption~\ref{assumption:SteinFormKernel} is satisfied.
    Observe that
    \[
        \EE_{\theta}(f,f)=-\E_{\theta}\left[\ang*{h_{yy}\left(X,\frac{\L p_{\theta}(X)}{p_{\theta}(X)}\right)\D_{\theta}f(X),\D_{\theta}f(X)}\right].
    \]
    Since $h_{yy}(x,y)$ is negative definite,
    $\EE_{\theta}(f,f)=0$ if and only if $\D_{\theta}f=0$, which is equivalent to $f$ being constant.
    Thus, Assumption~\ref{assumption:SteinFormKernel} is satisfied.
\end{proof}

\subsection{Existence and uniqueness of Stein score functions}
\label{sec:ExistenceUniquenessSteinScore}

\begin{definition}[Poincaré inequality]
    \label{def:PoincareInequality}
    The Dirichlet form $\EE_{\theta}$ defined in \eqref{eq:SteinForm} satisfies a Poincaré inequality with constant $\alpha(\theta)>0$
    if
    \[
    \EE_{\theta}(f,f)\geq \frac{1}{\alpha(\theta)}\Var_{\theta}[f(X)],
    \quad \forall f \in \operatorname{Dom}\A_{\theta}.
    \]
\end{definition}

A Poincaré inequality implies Assumption~\ref{assumption:SteinFormKernel}.
The following lemma specifies the kernel of $\A_{\theta}$. 

\begin{lemma}
    \label{lem:SteinOperatorKernel}
    For any $\theta\in\Theta$,
    under Assumption~\ref{assumption:SteinFormKernel}, $\A_{\theta}f=0$ holds if and only if $f$ is constant.
\end{lemma}

\begin{proof}
Suppose $\A_{\theta}f=0$. Then, by Lemma~\ref{lem:SteinFormSymmetry}, we have $\EE_{\theta}(f,f)=0$.
Thus, Assumption~\ref{assumption:SteinFormKernel} implies that $f$ is constant.
The converse follows immediately from $\D_{\theta}1=0$. 
\end{proof}

The following corollary states that the Stein score functions are unique under the kernel condition, which directly follows from Lemma~\ref{lem:SteinOperatorKernel}.
\begin{corollary}
    \label{cor:SteinScoreUnique}
    Suppose the Stein score functions $\psi_{\theta,1},\dots,\psi_{\theta,d}$ exist.
    Then, under Assumption~\ref{assumption:SteinFormKernel}, the Stein score functions are uniquely determined.
\end{corollary}

The following proposition gives a sufficient condition for the existence of Stein score functions based on a Poincaré inequality.
More generally, if $\A_{\theta}$ is the generator of a Markov process with invariant distribution $p_{\theta}$, the existence of the Stein score functions reduces to solving the Poisson equations
\[
    \A_{\theta}\psi_{\theta,j}=-\partial_{\theta_j}\log p_{\theta},\quad j=1,\dots,d.
\]
Alternative sufficient conditions for the solvability of Poisson equations include Foster--Lyapunov drift conditions; see, e.g., \citet{Glynn1996}.

\begin{proposition}
    Suppose that $\EE_{\theta}$ satisfies a Poincaré inequality with constant $\alpha(\theta)>0$ and that $\A_{\theta}$ is self-adjoint.
    Then, the Stein score functions $\psi_{\theta,1},\dots,\psi_{\theta,d}$ defined in Definition~\ref{def:SteinScore} exist at $\theta$.    
\end{proposition}

\begin{proof}
    Since $\A_{\theta}$ is self-adjoint and $\operatorname{Ker}\A_{\theta}=\R$ by Lemma~\ref{lem:SteinOperatorKernel}, we have
    \[
    \overline{\operatorname{Im}\A_{\theta}}=(\operatorname{Ker}\A_{\theta})^\perp=\{f\in L^2(p_{\theta})\mid \E_{\theta}[f(X)]=0\}.
    \]
    We show $\operatorname{Im}\A_{\theta}$ is closed. 
    By Lemma~\ref{lem:SteinFormSymmetry}, 
    for $f\in\operatorname{Dom}\A_{\theta}$ satisfying $\E_{\theta}[f(X)]=0$, we have
    \[
        -\E_{\theta}[f(X)\A_{\theta}f(X)]=\EE_{\theta}(f,f)\geq\frac{1}{\alpha(\theta)}\E_{\theta}[|f(X)|^2]
    \]
    Hence, by the Cauchy--Schwarz inequality,
    \begin{equation}
        \label{eq:proof_ExistenceSteinScore}
        \E_{\theta}[|\A_{\theta}f(X)|^2] \geq \frac{1}{\alpha(\theta)^2}\E_{\theta}[|f(X)|^2],
    \end{equation}
    Let $f^{(m)}\in\operatorname{Dom}\A_{\theta}$ be a sequence such that $\A_{\theta}f^{(m)}\to g$ in $L^2(p_{\theta})$.
    Define $\tilde{f}^{(m)}\coloneqq f^{(m)}-\E_{\theta}[f^{(m)}(X)]$.
    Since constant functions belong to $\operatorname{Ker}\A_{\theta}$, we have $\A_{\theta}\tilde{f}^{(m)}=\A_{\theta}f^{(m)}$.
    Moreover, $\E_{\theta}[\tilde{f}^{(m)}(X)]=0$.
    Applying \eqref{eq:proof_ExistenceSteinScore} to $\tilde{f}^{(m)}-\tilde{f}^{(n)}$ yields 
    \[
        \E_{\theta}[|\A_{\theta}f^{(m)}(X)-\A_{\theta}f^{(n)}(X)|^2]\geq\frac{1}{\alpha(\theta)^2}\E_{\theta}[|\tilde{f}^{(m)}(X)-\tilde{f}^{(n)}(X)|^2],
    \]
    Thus, $\tilde{f}^{(m)}$ is a Cauchy sequence in $L^2(p_{\theta})$.
    Hence, there exists $f\in L^2(p_{\theta})$ such that $\tilde{f}^{(m)}\to f$ in $L^2(p_{\theta})$.
    Since a self-adjoint operator is closed, we obtain $f\in\operatorname{Dom}\A_{\theta}$ and $\A_{\theta}f=g$.
    Hence, $\operatorname{Im}\A_{\theta}$ is closed.
    Consequently,
    \[
        \operatorname{Im}\A_{\theta}=\{f\in L^2(p_{\theta})\mid \E_{\theta}[f(X)]=0\}.
    \] 
    Since the Fisher score functions are centered and belong to $L^2(p_{\theta})$, $\partial_{\theta_j}\log p_{\theta}\in\operatorname{Im}\A_{\theta}$ for $j=1,\dots,d$.
    Hence, there exists $\psi_{\theta,j}\in\operatorname{Dom}\A_{\theta}$ satisfying Definition~\ref{def:SteinScore}, which completes the proof.
\end{proof}

\subsection{Auxiliary lemmas and proof of Theorem~\ref{thm:AsymptoticEfficiency}}
\label{sec:proof_AsymptoticEfficiency}

We first prove Lemma~\ref{lem:SteinFormSymmetry} and then establish the auxiliary results used in the proof of Theorem~\ref{thm:AsymptoticEfficiency}.

\begin{proof}[Proof of Lemma~\ref{lem:SteinFormSymmetry}]
   Observe that
   \begin{align*}
       \E_{\theta}[f(X)\A_{\theta}g(X)]
        &=\int_\X p_{\theta}(x)f(x)\L^\ast\left[ h_{yy}\left(\,\cdot\,,\frac{\L p_{\theta}(\,\cdot\,)}{p_{\theta}(\,\cdot\,)}\right)\D_{\theta}g \right](x)\d x\\
        &\phantom{=}-\int_\X p_{\theta}(x)f(x)\ang*{h_{yy}\left(x,\frac{\L p_{\theta}(x)}{p_{\theta}(x)}\right)\D_{\theta}g(x),\frac{\L p_{\theta}(x)}{p_{\theta}(x)}}\d x\\
        &=\int_\X \ang*{h_{yy}\left(x,\frac{\L p_{\theta}(x)}{p_{\theta}(x)}\right)\D_{\theta}g(x),\L[fp_{\theta}](x)}\d x\\
        &\phantom{=}-\int_\X p_{\theta}(x)\ang*{h_{yy}\left(x,\frac{\L p_{\theta}(x)}{p_{\theta}(x)}\right)\D_{\theta}g(x),\frac{f(x)\L p_{\theta}(x)}{p_{\theta}(x)}}\d x\\
        &=-\EE_{\theta}(f,g).
    \end{align*}
    Similarly, we have $\EE_{\theta}(f,g)=-\E_{\theta}[g(X)\A_{\theta}f(X)]$, which completes the proof.
\end{proof}

The following lemma relates $G(\theta)$ to the Dirichlet form $\EE_{\theta}$.

\begin{lemma}
    \label{lem:SteinFormMetric}
    Suppose the regularity condition R3 holds. Then, we have
    \[
    G(\theta^\star)_{jk}=\EE_{\theta^\star}(\partial_{\theta_j}\log p_{\theta^\star},\partial_{\theta_k}\log p_{\theta^\star}).
    \]
\end{lemma}

\begin{proof}
    By R3 and Lemma~\ref{lem:SteinFormSymmetry}, we have
    \begin{align*}
        G(\theta^\star)_{jk}
        &=\int_\X \partial_{\theta_k}(p_{\theta}(x)\A_{\theta}\partial_{\theta_j}\log p_{\theta}(x))|_{\theta=\theta^\star}\d x-\E_{\theta^\star}[\partial_{\theta_k}\log p_{\theta^\star}(X)\A_{\theta^\star}\partial_{\theta_j}\log p_{\theta^\star}(X)]\\
        &=\partial_{\theta_k}\E_{\theta}[\A_{\theta}\partial_{\theta_j}\log p_{\theta}(X)]|_{\theta=\theta^\star}+\EE_{\theta^\star}(\partial_{\theta_j}\log p_{\theta^\star},\partial_{\theta_k}\log p_{\theta^\star})
    \end{align*}
    By Stein's identity, we have $\E_{\theta}[\A_{\theta}\partial_{\theta_j}\log p_{\theta}(X)]=0$ for all $\theta\in\Theta$.
    Hence, 
    \[G(\theta^\star)_{jk}=\EE_{\theta^\star}(\partial_{\theta_j}\log p_{\theta^\star},\partial_{\theta_k}\log p_{\theta^\star}),
    \]
    which completes the proof.
\end{proof}

For $j=1,\dots,d$, define $u_{\theta^\star,j}\in\operatorname{Dom}\A_{\theta^\star}$ by
\[
u_{\theta^\star,j}\coloneqq \sum_{k=1}^{d}(G(\theta^\star)I(\theta^\star)^{-1})_{jk}\psi_{\theta^\star,k}-\partial_{\theta_{j}}\log p_{\theta^\star},\quad j=1,\dots,d.
\]
Lemma~\ref{lem:SteinFormMetric} implies that $u_{\theta^\star,1},\dots,u_{\theta^\star,d}$ are $\EE_{\theta^\star}$-orthogonal to $\partial_{\theta_j}\log p_{\theta^\star}$ for $j=1,\dots,d$, that is,
\[
\EE_{\theta^\star}(u_{\theta^\star,j},\partial_{\theta_k}\log p_{\theta^\star})=0,\quad j,k=1,\dots,d.
\] 

The following lemma gives an asymptotic variance decomposition.
\begin{lemma}
    \label{lem:AsymptoticVarianceGap}
    Under the regularity conditions R1-R5, we have
    \[
    \AVar[\hat\theta]-\AVar[\hat\theta_{\mathrm{MLE}}]=G(\theta^\star)^{-1}
    \E_{\theta^\star}\left[
    \begin{pmatrix}\A_{\theta^\star}u_{\theta^\star,1}(X)\\\vdots\\\A_{\theta^\star}u_{\theta^\star,d}(X)\end{pmatrix}
    \begin{pmatrix}\A_{\theta^\star}u_{\theta^\star,1}(X)\\\vdots\\\A_{\theta^\star}u_{\theta^\star,d}(X)\end{pmatrix}^\top
    \right]
    G(\theta^\star)^{-1}.
    \]
\end{lemma}

\begin{proof}
    The proof follows the argument used in the proof of Lemma~4 in \citet{Nagai2026}.
    $\hat\theta_\mathrm{MLE}$ has the following asymptotic linear representation:
    \begin{align*}
        \hat\theta_\mathrm{MLE}-\theta^\star
        &=I(\theta^\star)^{-1}\frac1n\sum_{i=1}^{n}\nabla_{\theta}\log p_{\theta^\star}(X_i)+\smallop[n^{-1/2}]\\
        &=-I(\theta^\star)^{-1}\frac1n\sum_{i=1}^{n}\begin{pmatrix}\A_{\theta^\star}\psi_{\theta^\star,1}(X_i)\\\vdots\\\A_{\theta^\star}\psi_{\theta^\star,d}(X_i)\end{pmatrix}+\smallop[n^{-1/2}]\\
        &=-G(\theta^\star)^{-1}\frac1n\sum_{i=1}^{n}\begin{pmatrix}\A_{\theta^\star}[\partial_{\theta_1}\log p_{\theta^\star}+u_{\theta^\star,1}](X_i)\\\vdots\\\A_{\theta^\star}[\partial_{\theta_d}\log p_{\theta^\star}+u_{\theta^\star,d}](X_i)\end{pmatrix}+\smallop[n^{-1/2}].
    \end{align*}

    Thus, we obtain the decomposition of the asymptotic variance:
    \[
        \AVar[\hat\theta_{\mathrm{MLE}}]=\AVar[\hat\theta]+G(\theta^\star)^{-1}(T+S+S^{\top})G(\theta^\star)^{-1},
    \]
    where $S,T\in\R^{d\times d}$  are the matrices whose $(j,k)$-th entries are defined by
    \[
        S_{jk}\coloneqq \E_{\theta^\star}[\A_{\theta^\star}u_{\theta^\star,j}(X)\A_{\theta^\star}\partial_{\theta_k}\log p_{\theta^\star}(X)],\enspace 
        T_{jk}\coloneqq \E_{\theta^\star}[\A_{\theta^\star}u_{\theta^\star,j}(X)\A_{\theta^\star}u_{\theta^\star,k}(X)].
    \]
    Observe the relation
    \begin{align*}
        S_{jk}
        &=\sum_{l=1}^{d}(G(\theta^\star)I(\theta^\star)^{-1})_{jl}\E_{\theta^\star}[\A_{\theta^\star}\psi_{\theta^\star,l}(X)\A_{\theta^\star}\partial_{\theta_{k}}\log p_{\theta^\star}(X)]\\
        &\phantom{=}-\E_{\theta^\star}[\A_{\theta^\star}\partial_{\theta_{j}}\log p_{\theta^\star}(X)\A_{\theta^\star}\partial_{\theta_{k}}\log p_{\theta^\star}(X)]\\
        &=(G(\theta^\star)I(\theta^\star)^{-1}G(\theta^\star)-G(\theta^\star)\AVar[\hat\theta]G(\theta^\star))_{jk}.
    \end{align*}

    Therefore, we have
    \[
        \AVar[\hat\theta_{\mathrm{MLE}}]=\AVar[\hat\theta]+G(\theta^\star)^{-1}TG(\theta^\star)^{-1}+2I(\theta^\star)^{-1}-2\AVar[\hat\theta].
    \]
    Since $\AVar[\hat\theta_\mathrm{MLE}]=I(\theta^\star)^{-1}$, we finally obtain $ \AVar[\hat\theta]-\AVar[\hat\theta_{\mathrm{MLE}}]=G(\theta^\star)^{-1}TG(\theta^\star)^{-1}$, which completes the proof.
\end{proof}

\begin{proof}[Proof of Theorem~\ref{thm:AsymptoticEfficiency}]
    $(1)\Rightarrow(2)$:

    Suppose $\AVar[\hat\theta]=\AVar[\hat\theta_\mathrm{MLE}]$.
    By Lemma~\ref{lem:AsymptoticVarianceGap}, $\A_{\theta^\star}u_{\theta^\star,j}=0$ for $j=1,\dots,d$.
    Lemma~\ref{lem:SteinOperatorKernel} implies that $u_{\theta^\star,j}$ is constant.
    Since $\E_{\theta^\star}[u_{\theta^\star,j}(X)]=0$, we have $u_{\theta^\star,j}=0$.
    Thus, we obtain
    \[
        \partial_{\theta_{j}}\log p_{\theta^\star}=\sum_{k=1}^{d}(G(\theta^\star)I(\theta^\star)^{-1})_{jk}\psi_{\theta^\star,k},
    \]
    which completes $(1)\Rightarrow(2)$.

    $(2)\Rightarrow(1)$:
    Suppose there exists a nonsingular matrix $\Lambda(\theta^\star)\in\R^{d\times d}$ such that $\partial_{\theta_j}\log p_{\theta^\star}=\sum_{k=1}^d\Lambda(\theta^\star)_{jk}\psi_{\theta^\star,k}$ for $j=1,\dots,d$.
    By Lemma~\ref{lem:SteinFormMetric},
    \begin{align*}    
        G(\theta^\star)_{jk}
        &=\EE_{\theta^\star}(\partial_{\theta_j}\log p_{\theta^\star},\partial_{\theta_k}\log p_{\theta^\star})\\
        &=-\E_{\theta^\star}[\partial_{\theta_k}\log p_{\theta^\star}(X)\A_{\theta^\star}\partial_{\theta_j}\log p_{\theta^\star}(X)]\\
        &=\E_{\theta^\star}\left[\partial_{\theta_k}\log p_{\theta^\star}(X)\sum_{l=1}^{d}\Lambda(\theta^\star)_{jl}\partial_{\theta_l}\log p_{\theta^\star}(X)\right]\\
        &=(\Lambda(\theta^\star)I(\theta^\star))_{jk}.
    \end{align*}
    Therefore, 
    \begin{align*}
        \hat\theta-\theta^\star
        &=-G(\theta^\star)^{-1}\frac1n\sum_{i=1}^{n}\begin{pmatrix}\A_{\theta^\star}\partial_{\theta_1}\log p_{\theta^\star}(X_i)\\\vdots\\\A_{\theta^\star}\partial_{\theta_d}\log p_{\theta^\star}(X_i)\end{pmatrix}+\smallop[n^{-1/2}]\\
        &=I(\theta^\star)^{-1}\Lambda(\theta^\star)^{-1}\frac1n\sum_{i=1}^{n}\Lambda(\theta^\star)\nabla_{\theta}\log p_{\theta^\star}(X_i)+\smallop[n^{-1/2}]\\
        &=\hat\theta_\mathrm{MLE}-\theta^\star+\smallop[n^{-1/2}],
    \end{align*}
    which implies $(2)\Rightarrow(1)$.
\end{proof}

\section{Further examples}
\label{sec:FurtherExamples}

In this section, we provide further examples.

\subsection{Weighted gradient and diffusion score matching}

Let $\X\subset\R^p$ be a connected open set with a smooth boundary
and let $M\colon\X\to\R^{p\times p}$ be a smooth function such that $M(x)$ is positive definite, and thus $M(x)^{1/2}$ exists.
Define $\L f\coloneqq M^{1/2}\nabla_xf$.
$\L^\ast$ is given by
\[
    \L^\ast V\coloneqq-\div_x\left(M^{1/2}V\right).
\]
When we choose $h(x,y)\coloneqq -\norm{y}^2/2$, the corresponding scoring rule $S$, Stein operator $\A_{\theta}$, divergence $D(p_{\theta^\star}\,\|\,p_{\theta})$, and Dirichlet form $\EE_{\theta}$ are given as follows:
\begin{align*}
    S(x,p_{\theta})&=\div_x\left(M\nabla_x\log p_{\theta}\right)(x)+\frac12\norm*{M(x)^{1/2}\nabla_x \log p_{\theta}(x)}^{2},\\
    \A_{\theta}f(x)&=\div_x(M\nabla_x f)(x)+\ang{M(x)\nabla_x f(x),\nabla_x \log p_{\theta}(x)},\\
    D(p_{\theta^\star}\,\|\,p_{\theta})&=\frac12\E_{\theta^\star}\left[\norm*{M(X)^{1/2}(\nabla_x \log p_{\theta^\star}(X)-\nabla_x \log p_{\theta}(X))}^2\right],\\
    \EE_{\theta}(f,g)&=\E_{\theta}[\ang{M(X)\nabla_x f(X),\nabla_x g(X)}].
\end{align*}
This scoring rule corresponds to diffusion score matching \citep{Barp2019,Qin2024}.
It recovers weighted score matching \citep{Hyvarinen2007,Yu2019,Liu2022} when $M(x)$ is diagonal.
An appropriate choice of $M$ can weaken the boundary conditions required of $\P$.

The following proposition shows that Assumption~\ref{assumption:DKernel} is satisfied by choosing $\L f\coloneqq M^{1/2}\nabla_xf$.
\begin{proposition}
    \label{prp:DKernelWeightedGradient}
    Let $\L f\coloneqq M^{1/2}\nabla_xf$.
    Then, Assumption~\ref{assumption:DKernel} is satisfied.
\end{proposition}
\begin{proof}
    Fix an arbitrary $\theta\in\Theta$.
    Observe that
    \[
        \D_{\theta}f(x)=M(x)^{1/2}\nabla_x f(x).
    \]
    Since $M(x)$ is positive definite, $\D_{\theta}f=0$ if and only if $f$ is constant, which completes the proof.
\end{proof}

The following proposition states an alternative sufficient condition for $(\A_{\theta},\F_{\theta})$ to be a Stein pair.
\begin{proposition}
    \label{prp:SteinFormContractionWeightedGradient}
    Let $\L f\coloneqq M^{1/2}\nabla_xf,\L^\ast V\coloneqq-\div_x(M^{1/2}V)$ and suppose $h_{yy}(x,y)$ is negative definite.
    Suppose further that there exists a smooth function $\Psi\colon(0,\infty)\to\R$ such that $\Psi'>0$ and $\Psi(p_{\theta^\star}(\,\cdot\,)/p_{\theta}(\,\cdot\,))\in\F_{\theta}$ for all $\theta^\star,\theta\in\Theta$.
    Then $(\A_{\theta},\F_{\theta})$ is a Stein pair.
\end{proposition}
\begin{proof}
    Fix an arbitrary $\theta^\star\in\Theta$.
    We show that for $f_{\theta}\coloneqq \Psi(p_{\theta^\star}(\,\cdot\,)/p_{\theta}(\,\cdot\,))\in\F_{\theta}$,
    $\E_{\theta^\star}[\A_{\theta}f_{\theta}(X)]=0$ implies $\theta=\theta^\star$.
    Observe that
    \[
    \D_{\theta}f_{\theta}(x)=M(x)^{1/2}\nabla_xf_{\theta}(x)=\Psi'\left(\frac{p_{\theta^\star}(x)}{p_{\theta}(x)}\right)M(x)^{1/2}\nabla_x\frac{p_{\theta^\star}(x)}{p_{\theta}(x)}
    \]
    Thus, we have
    \begin{align*}    
    \E_{\theta^\star}\left[\A_{\theta}f_{\theta}(X)\right]
        &=\E_{\theta^\star}\left[\ang*{h_{yy}\left(X,\frac{\L p_{\theta}(X)}{p_{\theta}(X)}\right)\D_{\theta}f_{\theta}(X),M(X)^{1/2}\nabla_x\log\frac{p_{\theta^\star}(X)}{p_{\theta}(X)}}\right]\\
        &=\E_{\theta^\star}\left[\Psi'\left(\frac{p_{\theta^\star}(X)}{p_{\theta}(X)}\right)\frac{p_{\theta^\star}(X)}{p_{\theta}(X)}\right.\\
        &\left.\phantom{=}\times\ang*{h_{yy}\left(X,\frac{\L p_{\theta}(X)}{p_{\theta}(X)}\right)M(X)^{1/2}\nabla_x\log\frac{p_{\theta^\star}(X)}{p_{\theta}(X)},M(X)^{1/2}\nabla_x\log\frac{p_{\theta^\star}(X)}{p_{\theta}(X)}}\right].
    \end{align*}
    Since $\Psi'(p_{\theta^\star}(x)/p_{\theta}(x))>0$ and $M(x)^{1/2}h_{yy}(x,y)M(x)^{1/2}$ is negative definite,
    \[
        \E_{\theta^\star}[\A_{\theta}f_{\theta}(X)]=0 \iff \nabla_x \log \frac{p_{\theta^\star}}{p_{\theta}}=0,
    \]
    which is equivalent to $\theta=\theta^\star$.
    Together with part (2) of Theorem~\ref{thm:ScoringRuleStein}, this shows that $(\A_{\theta},\F_{\theta})$ is a Stein pair, which completes the proof.
\end{proof}

\begin{example}[Gamma distribution]
    The gamma distribution with a fixed shape parameter $\alpha>0$ and rate parameter $\theta>0$ has the density
    \[
        p_{\theta}(x)=\frac{\theta^{\alpha}}{\Gamma(\alpha)}x^{\alpha-1}e^{-x\theta},\enspace x>0.
    \]
    Let $M(x)=x$ and $h(x,y)=-y^2/2$.
    Then the corresponding Stein operator is given by
    \[
        \A_{\theta}f(x)
        =(\alpha-\theta x)f'(x)+xf''(x).
    \]
    Observe that
    \[
      \A_{\theta}\partial_{\theta}\log p_{\theta}(x)=-\theta\partial_{\theta}\log p_{\theta}(x).
    \]
    Thus, the Stein score is given by $\psi_{\theta}=\theta^{-1}\partial_{\theta}\log p_{\theta}$, which satisfies part (2) of Theorem~\ref{thm:AsymptoticEfficiency}.
\end{example}

\subsection{Local scoring rules on discrete spaces}
\label{sec:ExamplesLocalScoringRuleDiscrete}

Let $(\X,E)$ be an undirected graph with finite vertex set $\X$ and edge set $E$.
For each $x\in\X$, define the set of neighbors of $x$ by
\[
    N(x)\coloneqq\{v\in \X\mid \{x,v\}\in E\}.
\]

Let
\begin{equation}
    \label{eq:NeighborhoodOperator}
    \L f(x)\coloneqq(1_{N(x)}(v)f(v))_{v\in\X},
\end{equation}
where $1_{N(x)}$ denotes the indicator function of $N(x)$. 
$\L^\ast$ is given by
\begin{equation}
    \label{eq:NeigborhoodOperatorAdjoint}    
    \L^\ast V(x)\coloneqq\sum_{v\in N(x)}V(v)_{x}.
\end{equation}

Note that the setting in Section~\ref{sec:ExamplesRatioMatching} is recovered by taking $\X\coloneqq\{\pm1\}^{p}$ and $N(x)\coloneqq\{x^{(j)}\mid j=1,\dots,p\}$,
whereas the setting in Section~\ref{sec:ExamplesPseudoLikelihood} is recovered by taking $\X\coloneqq\X_1\times\dots\times\X_p$ and $N(x)\coloneqq\{v\in\X\setminus\{x\}\mid \exists j=1,\dots,p,\enspace v_{-j}=x_{-j}\}$.

Let $\phi\colon[0,\infty)\to\R$ be a smooth concave function such that $\phi(0)=0$.
When we choose $h(x,y)\coloneqq \sum_{v\in\X}\phi(y_v)$, the corresponding scoring rule $S$, Stein operator $\A_{\theta}$, divergence $D(p_{\theta^\star}\,\|\,p_{\theta})$, and Dirichlet form $\EE_{\theta}$ are given as follows:
\begin{align}
    S(x,p_{\theta})&= \sum_{v\in N(x)}\left(\phi\left(\frac{p_{\theta}(v)}{p_{\theta}(x)}\right)+\phi'\left(\frac{p_{\theta}(x)}{p_{\theta}(v)}\right)-\phi'\left(\frac{p_{\theta}(v)}{p_{\theta}(x)}\right)\frac{p_{\theta}(v)}{p_{\theta}(x)}\right),\notag\\
    \A_{\theta}f(x)&=
    -\sum_{v\in N(x)}\left( \phi''\left(\frac{p_{\theta}(x)}{p_{\theta}(v)}\right)\frac{p_{\theta}(x)}{p_{\theta}(v)}+\phi''\left(\frac{p_{\theta}(v)}{p_{\theta}(x)}\right)\frac{p_{\theta}(v)^2}{p_{\theta}(x)^2} \right)(f(v)-f(x)),\label{eq:NeighborhoodSteinOperator}\\
    D(p_{\theta^\star}\,\|\,p_{\theta})
    &=\sum_{v\in\X}\E_{\theta^\star}\left[1_{N(X)}(v)\left(\phi\left(\frac{p_{\theta}(v)}{p_{\theta}(X)}\right)-\phi\left(\frac{p_{\theta^{\star}}(v)}{p_{\theta^{\star}}(X)}\right)\right.\right.\notag\\
    &\phantom{=}\left.\left.-\phi'\left(\frac{p_{\theta}(v)}{p_{\theta}(X)}\right)\left(\frac{p_{\theta}(v)}{p_{\theta}(X)}-\frac{p_{\theta^{\star}}(v)}{p_{\theta^{\star}}(X)}\right)\right)\right],\notag\\
    \EE_{\theta}(f,g)
    &=-\sum_{v\in\X}\E_{\theta}\left[ \phi''\left(\frac{p_{\theta}(v)}{p_{\theta}(X)}\right)\frac{p_{\theta}(v)^2}{p_{\theta}(X)^2}
    1_{N(X)}(v)(f(v)-f(X))(g(v)-g(X))\right].\notag
\end{align}

This scoring rule is local in the sense of \citet{Dawid2012}.
\citet{Kanamori2017} studied conditions for the strict propriety of graph-based local scoring rules in terms of the underlying neighborhood system and provided various examples.
Independently of the literature on local scoring rules, \citet{Meng2022} developed a closely related graph-based formulation of score matching for discrete data.

\begin{remark}[Comparison with \citet{Yang2018}]
    \citet{Yang2018} characterized Stein operators on discrete spaces satisfying Stein's identity as operators of the form
    \[
    \A_{\theta}f(x)=\sum_{v\in\X}\left(\frac{p_{\theta}(v)}{p_{\theta}(x)}g_{\theta}(x,v)f(x)-g_{\theta}(v,x)f(v)\right),
    \]
    where $g_{\theta}\colon\X\times\X\to\R$.
    The Stein operator defined in \eqref{eq:NeighborhoodSteinOperator} is recovered by choosing
    \[
        g_{\theta}(x,v)\coloneqq 1_{N(x)}(v)
        \left( \phi''\left(\frac{p_{\theta}(v)}{p_{\theta}(x)}\right)\frac{p_{\theta}(v)}{p_{\theta}(x)}+\phi''\left(\frac{p_{\theta}(x)}{p_{\theta}(v)}\right)\frac{p_{\theta}(x)^2}{p_{\theta}(v)^2}\right).
    \]
\end{remark}

The following proposition shows that Assumption~\ref{assumption:DKernel} is satisfied in this setting.
\begin{proposition}
    \label{prp:DKernelLocalScoringRuleDiscrete}
    Let $\L$ be defined in \eqref{eq:NeighborhoodOperator}.
    Suppose $(\X,E)$ is connected.
    Then, Assumption~\ref{assumption:DKernel} is satisfied.
\end{proposition}

\begin{proof}
    Fix an arbitrary $\theta\in\Theta$.
    Observe that
    \[
        \D_{\theta}f(x)=\left(\frac{p_{\theta}(v)}{p_{\theta}(x)}(f(v)-f(x))1_{N(x)}(v)\right)_{v\in\X}.
    \]
    Thus, $\D_{\theta}f(x)=0$ is equivalent to $f(v)=f(x)$ for all $x\in\X$ and $v\in N(x)$.
    Since $(\X,E)$ is connected, this is equivalent to $f$ being constant, which completes the proof.
\end{proof}

The following proposition states an alternative sufficient condition for $(\A_{\theta},\F_{\theta})$ to be a Stein pair.
\begin{proposition}
    \label{prp:SteinFormContractionLocalScoringRuleDiscrete}
    Let $\L$ and $\L^\ast$ be defined in \eqref{eq:NeighborhoodOperator} and \eqref{eq:NeigborhoodOperatorAdjoint}, respectively.
    Suppose $(\X,E)$ is connected and $h_{yy}(x,y)$ is negative definite and diagonal.
    Suppose further that there exists a strictly increasing function $\Psi\colon(0,\infty)\to\R$ such that $\Psi(p_{\theta^\star}(\,\cdot\,)/p_{\theta}(\,\cdot\,))\in\F_{\theta}$ for all $\theta^\star,\theta\in\Theta$.
    Then $(\A_{\theta},\F_{\theta})$ is a Stein pair.
\end{proposition}

\begin{proof}
    Fix $\theta^\star\in\Theta$. We show that for $f_{\theta}\coloneqq\Psi(p_{\theta^\star}(\,\cdot\,)/p_{\theta}(\,\cdot\,))\in\F_{\theta}$,
    $\E_{\theta^\star}[\A_{\theta}f_{\theta}(X)]=0$ implies $\theta=\theta^\star$.
    Observe the relation
    \begin{align*}
        \E_{\theta^\star}[\A_{\theta}f_{\theta}(X)]
        &=\sum_{v\in\X}\E_{\theta}\left[ h_{yy}\left(X,\frac{\L p_{\theta}(X)}{p_{\theta}(X)}\right)_{vv}\frac{p_{\theta}(v)^2}{p_{\theta}(X)^2}1_{N(X)}(v)\right.\\
        &\phantom{=}\times\left.\left(\Psi\left(\frac{p_{\theta^\star}(v)}{p_{\theta}(v)}\right)-\Psi\left(\frac{p_{\theta^\star}(X)}{p_{\theta}(X)}\right)\right)\left(\frac{p_{\theta^\star}(v)}{p_{\theta}(v)}-\frac{p_{\theta^\star}(X)}{p_{\theta}(X)}\right)\right].
    \end{align*}
    Since $h_{yy}(x,y)$ is negative definite and $\Psi$ is strictly increasing,
    $\E_{\theta^\star}[\A_{\theta}f_{\theta}(X)]=0$ holds if and only if $1_{N(x)}(v)(p_{\theta^\star}(v)/p_{\theta}(v)-p_{\theta^\star}(x)/p_{\theta}(x))=0$ for all $x,v\in\X$.
    Since $(\X,E)$ is connected, this is equivalent to $p_{\theta^\star}/p_{\theta}$ being constant, that is, $\theta=\theta^\star$, which completes the proof. 
\end{proof}

\begin{example}[Pseudo-spherical score and $\gamma$-divergence]
    Let $(\X,E)$ be a complete graph with self-loops, that is, $N(x)=\X$ for all $x\in\X$.
    In this setting, the pseudo-spherical scoring rule \citep{Fujisawa2008,Kanamori2015} is recovered by choosing
    \[
      h(x,y)\coloneqq-\frac{1}{|\X|\gamma}\left(\sum_{v\in\X}y_{v}^{1+\gamma}\right)^{1/(1+\gamma)},
    \]    
    where $\gamma>0$.
    The associated divergence is equivalent to the $\gamma$-divergence up to a strictly monotone transformation.
    The Stein operator induced by the pseudo-spherical scoring rule is
    \[
        \A_{\theta}f(x)=\frac{p_{\theta}(x)^{\gamma}}{\E_{\theta}[p_{\theta}(X)^{\gamma}]^{\gamma/(1+\gamma)}}\left(\frac{\E_{\theta}[f(X)p_{\theta}(X)^{\gamma}]}{\E_{\theta}[p_{\theta}(X)^{\gamma}]}-f(x)\right).
    \]
    Although the pseudo-spherical score is closely related to the $\gamma$-divergence, the above Stein operator is distinct from the $\gamma$-Stein operator of \citet{Eguchi2026}, which is derived from the first variation of the $\gamma$-divergence along infinitesimal transports on the sample space.
\end{example}

\subsection{Shift operator}

Let $\X\subset\mathbb{Z}$ and $\L f(x)\coloneqq f(x+1)1_{\X}(x+1)$.
$\L^\ast$ is given by $\L^\ast f(x)\coloneqq f(x-1)1_{\X}(x-1)$.

When we choose $h(x,y)\coloneqq -w(x)y^2/2$, where $w(x)>0$, the corresponding scoring rule $S$, Stein operator $\A_{\theta}$, divergence $D(p_{\theta^\star}\,\|\,p_{\theta})$, and Dirichlet form $\EE_{\theta}$ are given as follows:
\begin{align}
    S(x,p_{\theta})&=\frac12w(x)1_{\X}(x+1)\frac{p_{\theta}(x+1)^{2}}{p_{\theta}(x)^{2}}-1_{\X}(x-1)w(x-1)\frac{p_{\theta}(x)}{p_{\theta}(x-1)},\notag\\
    \A_{\theta}f(x)&=-1_{\X}(x-1)w(x-1)\frac{p_{\theta}(x)}{p_{\theta}(x-1)}(f(x)-f(x-1))\notag\\
    &\phantom{=}+1_{\X}(x+1)w(x)\frac{p_{\theta}(x+1)^{2}}{p_{\theta}(x)^{2}}(f(x+1)-f(x)),\label{eq:ShiftSteinOperator}\\
    D(p_{\theta^\star}\,\|\,p_{\theta})&=\frac12\E_{\theta^{\star}}\left[w(X)1_{\X}(X+1)\left(\frac{p_{\theta}(X+1)}{p_{\theta}(X)}-\frac{p_{\theta^\star}(X+1)}{p_{\theta^\star}(X)}\right)^{2}\right],\notag\\
    \EE_{\theta}(f,g)&=\E_{\theta}\left[w(X)1_{\X}(X+1)\frac{p_{\theta}(X+1)^{2}}{p_{\theta}(X)^{2}}(f(X+1)-f(X))(g(X+1)-g(X))\right]\notag.
\end{align}

\begin{remark}[Comparison with \citet{Fischer2025discrete}]
    \citet{Fischer2025discrete} constructed closed-form estimators for discrete distributions using the following Stein operator:
    \[
    \mathcal{T}_{\theta}f(x)=\frac{1_{\X}(x+1)f(x+1)p_{\theta}(x+1)\tau_{\theta}(x+1)}{p_{\theta}(x)}-f(x)\tau_{\theta}(x).
    \]
    Choosing $\tau_{\theta}(x)=1_{\X}(x-1)w(x-1)$, the Stein operator defined in \eqref{eq:ShiftSteinOperator} can be written as
    \[
        \A_{\theta}f(x)=
        \mathcal{T}_{\theta}g_{\theta}(x),\quad 
        g_{\theta}(x)\coloneqq\frac{1_{\X}(x-1)p_{\theta}(x)}{p_{\theta}(x-1)}(f(x)-f(x-1)).
    \]
\end{remark}

The following proposition shows that Assumption~\ref{assumption:DKernel} is satisfied in this setting.
\begin{proposition}
    \label{prp:DKernelShift}
    Let $\L f(x)\coloneqq f(x+1)1_\X(x+1)$ and suppose $\X$ is an integer interval.
    Then, Assumption~\ref{assumption:DKernel} is satisfied.
\end{proposition}

\begin{proof}
    Fix an arbitrary $\theta\in\Theta$.
    Observe that
    \[
        \D_{\theta}f(x)=\frac{p_{\theta}(x+1)}{p_{\theta}(x)}(f(x+1)-f(x))1_{\X}(x+1).
    \]
    Thus, $\D_{\theta}f(x)=0$ is equivalent to $f(x+1)=f(x)$ for every $x\in\X$ such that $x+1\in\X$.
    Since $\X$ is an integer interval, this is equivalent to $f$ being constant, which completes the proof.
\end{proof}

The following proposition states an alternative sufficient condition for $(\A_{\theta},\F_{\theta})$ to be a Stein pair.
\begin{proposition}
    \label{prp:SteinFormContractionShift}
    Let $\L f(x)\coloneqq f(x+1)1_\X(x+1)$ and $\L^\ast f(x)\coloneqq f(x-1)1_\X(x-1)$.
    Suppose $\X$ is an integer interval and $h_{yy}(x,y)<0$.
    Suppose further that there exists a strictly increasing function $\Psi\colon(0,\infty)\to\R$ such that $\Psi(p_{\theta^\star}(\,\cdot\,)/p_{\theta}(\,\cdot\,))\in\F_{\theta}$ for all $\theta^\star,\theta\in\Theta$.
    Then $(\A_{\theta},\F_{\theta})$ is a Stein pair.
\end{proposition}

\begin{proof}
    Fix $\theta^\star\in\Theta$. We show that for $f_{\theta}\coloneqq\Psi(p_{\theta^\star}(\,\cdot\,)/p_{\theta}(\,\cdot\,))\in\F_{\theta}$,
    $\E_{\theta^\star}[\A_{\theta}f_{\theta}(X)]=0$ implies $\theta=\theta^\star$.
    Observe the relation
    \begin{align*}
        \E_{\theta^\star}[\A_{\theta}f_{\theta}(X)]
        &=\E_{\theta}\left[ h_{yy}\left(X,\frac{\L p_{\theta}(X)}{p_{\theta}(X)}\right)\frac{p_{\theta}(X+1)^2}{p_{\theta}(X)^2}1_{\X}(X+1)\right.\\
        &\phantom{=}\times\left.\left(\Psi\left(\frac{p_{\theta^\star}(X+1)}{p_{\theta}(X+1)}\right)-\Psi\left(\frac{p_{\theta^\star}(X)}{p_{\theta}(X)}\right)\right)\left(\frac{p_{\theta^\star}(X+1)}{p_{\theta}(X+1)}-\frac{p_{\theta^\star}(X)}{p_{\theta}(X)}\right)\right].
    \end{align*}
    Since $h_{yy}(x,y)<0$ and $\Psi$ is strictly increasing,
    $\E_{\theta^\star}[\A_{\theta}f_{\theta}(X)]=0$ holds if and only if $1_{\X}(x+1)(p_{\theta^\star}(x+1)/p_{\theta}(x+1)-p_{\theta^\star}(x)/p_{\theta}(x))=0$ for all $x\in\X$.
    Since $\X$ is an integer interval, this is equivalent to $p_{\theta^\star}/p_{\theta}$ being constant, that is, $\theta=\theta^\star$, which completes the proof. 
\end{proof}

\begin{example}[Poisson distribution]
    The Poisson distribution with rate parameter $\theta>0$ has the density
    \[
        p_{\theta}(x)=\frac{\theta^{x}}{x!}e^{-\theta},\quad x\in\{0,1,\dots\}.
    \]
    Let $h(x,y)=-(x+1)^2y^2/2$.
    Then the corresponding Stein operator is given by
    \[
        \A_{\theta}f(x)=-\theta x(f(x)-f(x-1))+\theta^{2}(f(x+1)-f(x))
    \]
    Observe that
    \[
        \A_{\theta}\partial_{\theta}\log p_{\theta}(x)
        =-\theta\partial_{\theta}\log p_{\theta}(x).
    \]
    Thus, the Stein score is given by $\psi_{\theta}=\theta^{-1}\partial_{\theta}\log p_{\theta}$, which satisfies part (2) of Theorem~\ref{thm:AsymptoticEfficiency}.
\end{example}

\subsection{Weighted integral operator}
\label{sec:ExamplesWeightedIntegral}

Let $\alpha,\beta\in L^1(\P)$ be strictly positive functions.
Define
\begin{equation}
    \label{eq:WeightedIntegralOperator}
    \L f(x)\coloneqq \alpha(x)\I_{\beta}[f],
\end{equation}
where $\I_{\beta}$ is the integral operator defined by
\[
\I_{\beta}[f]\coloneqq \int_{\X}f(x)\beta(x)\d x.
\]
$\I_{\alpha}$ is defined similarly.
$\L^\ast$ is given by
\begin{equation}
    \label{eq:WeightedIntegralOperatorAdjoint}    
    \L^\ast f(x)\coloneqq \beta(x)\I_{\alpha}[f].
\end{equation}

In this setting, the results in Section~\ref{sec:ExamplesNCE} are recovered by choosing $\alpha(x)=n(x),\beta(x)=1$, while those in Section~\ref{sec:ExamplesMLE} are recovered by choosing $\alpha(x)=\beta(x)=1$.

The following proposition shows that Assumption~\ref{assumption:DKernel} is satisfied in this setting.
\begin{proposition}
    \label{prp:DKernelWeightedIntegralOperator}
    Let $\L f(x)\coloneqq\alpha(x)\I_{\beta}[f]$.
    Then, Assumption~\ref{assumption:DKernel} is satisfied.
\end{proposition}

\begin{proof}
    Fix an arbitrary $\theta\in\Theta$.
    Observe that
    \[
        \D_{\theta}f(x)=\frac{\alpha(x)\I_{\beta}[p_{\theta}]}{p_{\theta}(x)}\left(\frac{\I_{\beta}[fp_{\theta}]}{\I_{\beta}[p_{\theta}]}-f(x)\right).
    \]
    Thus, $\D_{\theta}f=0$ if and only if $f$ is constant, which completes the proof.
\end{proof}

The following proposition states an alternative sufficient condition for $(\A_{\theta},\F_{\theta})$ to be a Stein pair.
\begin{proposition}
    \label{prp:SteinFormContractionWeightedIntegral}
    Let $\L$ and $\L^\ast$ be defined in \eqref{eq:WeightedIntegralOperator} and \eqref{eq:WeightedIntegralOperatorAdjoint}, respectively.
    Suppose $h_{yy}(x,y)<0$.
    Suppose further that $1_{\{p_{\theta^\star}/p_{\theta}>\I_\beta[p_{\theta^\star}]/\I_\beta[p_\theta]\}}(\,\cdot\,)\in\F_{\theta}$ for all $\theta^\star,\theta\in\Theta$.
    Then $(\A_{\theta},\F_{\theta})$ is a Stein pair.
\end{proposition}

\begin{proof}
    Fix $\theta^\star\in\Theta$. 
    Let 
    \[
        r_{\theta}(x)\coloneqq\frac{p_{\theta^\star}(x)}{p_{\theta}(x)}-\frac{\I_\beta[p_{\theta^\star}]}{\I_\beta[p_\theta]}.
    \]
    We show that for $f_{\theta}\coloneqq1_{\{r_\theta>0\}}(\,\cdot\,)\in\F_{\theta}$,
    $\E_{\theta^\star}[\A_{\theta}f_{\theta}(X)]=0$ implies $\theta=\theta^\star$.
    Since
    \[
    \I_\beta[r_{\theta}p_{\theta}]=\I_\beta[p_{\theta^\star}]-\I_\beta[p_{\theta^\star}]=0,
    \]
    $r_\theta$ cannot be nonnegative or nonpositive unless it is identically zero.
    Thus, if $r_{\theta}$ is not identically zero, it takes both positive and negative values on sets of positive measure, and hence
    \begin{equation}
        \label{eq:SteinFormContractionWeightedIntegral}
        0<\frac{\I_\beta[f_\theta p_\theta]}{\I_\beta[p_\theta]}<1.
    \end{equation}
    Observe the relation
    \begin{align*}
        \E_{\theta^\star}[\A_{\theta}f_{\theta}(X)]
        =\E_{\theta}\left[h_{yy}\left(X,\frac{\alpha(X)\I_{\beta}[p_{\theta}]}{p_{\theta}(X)}\right)\left(\frac{\alpha(X)\I_\beta[p_{\theta}]}{p_{\theta}(X)}\right)^{2}
        \left(f_{\theta}(X)-\frac{\I_{\beta}[f_{\theta}p_{\theta}]}{\I_{\beta}[p_{\theta}]}\right)
        r_{\theta}(X)\right].
    \end{align*}
    By \eqref{eq:SteinFormContractionWeightedIntegral},
    \[
       \left(f_{\theta}(x)-\frac{\I_{\beta}[f_{\theta}p_{\theta}]}{\I_{\beta}[p_{\theta}]}\right)r_\theta(x)\geq0
    \]
    for all $x\in\X$, and equality holds if and only if $r_{\theta}(x)=0$.
    Since $h_{yy}(x,y)<0$, the preceding identity implies that $\E_{\theta^\star}[\A_{\theta}f_{\theta}(X)]=0$ holds if and only if $r_{\theta}=0$, that is, $\theta=\theta^\star$, which completes the proof. 
\end{proof}

The following proposition provides a closed-form expression for the Stein score functions.

\begin{proposition}
    \label{prp:SteinScoreWeightedIntegral}
    Suppose $h_{yy}(x,y)<0$.
    For $j=1,\dots,d$, define
    \[
    \tilde{\psi}_{\theta,j}(x)\coloneqq\frac{C_{\theta,j}\beta(x)-\partial_{\theta_j}\log p_{\theta}(x)}{h_{yy}(x,\alpha(x)\I_{\beta}[p_{\theta}]/p_{\theta}(x))}\left(\frac{p_{\theta}(x)}{\alpha(x)\I_{\beta}[p_{\theta}]}\right)^2,
    \]
    where $C_{\theta,j}$ is a constant defined by
    \[
        C_{\theta,j}\coloneqq
        \I_{\beta}\left[\frac{\partial_{\theta_j}p_{\theta}}{h_{yy}(\,\cdot\,,\alpha(\,\cdot\,)\I_{\beta}[p_{\theta}]/p_{\theta}(\,\cdot\,))}\left(\frac{p_{\theta}}{\alpha\I_{\beta}[p_{\theta}]}\right)^2\right]
        \I_{\beta}\left[\frac{p_{\theta}\beta}{h_{yy}(\,\cdot\,,\alpha(\,\cdot\,)\I_\beta[p_{\theta}]/p_{\theta}(\,\cdot\,))}\left(\frac{p_{\theta}}{\alpha\I_{\beta}[p_{\theta}]}\right)^2\right]^{-1}.
    \]
    Then, the Stein score functions are given by
    \[
        \psi_{\theta,j}(x)=\tilde{\psi}_{\theta,j}(x)-\E_{\theta}[\tilde{\psi}_{\theta,j}(X)],\enspace j=1,\dots,d.
    \]
    Here, we assume that the quantities $C_{\theta,j}$ are well defined and that $\psi_{\theta,j}\in\operatorname{Dom}\A_{\theta}$ for $j=1,\dots,d$.    
\end{proposition}

\begin{proof}
    Observe that 
    \begin{align*}
        \I_{\beta}[\tilde{\psi}_{\theta,j}p_{\theta}]
        &=C_{\theta,j}\I_{\beta}\left[
            \frac{p_{\theta}\beta}{h_{yy}(\,\cdot\,,\alpha(\,\cdot\,)\I_{\beta}[p_{\theta}]/p_{\theta}(\,\cdot\,))}\left(\frac{p_{\theta}}{\alpha\I_{\beta}[p_{\theta}]}\right)^{2}
        \right]\\
        &\phantom{=}-\I_{\beta}\left[\frac{\partial_{\theta_j}p_{\theta}}{h_{yy}(\,\cdot\,,\alpha(\,\cdot\,)\I_{\beta}[p_{\theta}]/p_{\theta}(\,\cdot\,))}\left(\frac{p_{\theta}}{\alpha\I_{\beta}[p_{\theta}]}\right)^2\right]\\
        &=0.
    \end{align*}
    
    The Stein operator $\A_{\theta}$ is given by
    \begin{align*}        
        \A_{\theta}f(x)
        &=\beta(x)\I_{\alpha}\left[h_{yy}\left(\,\cdot\,,\frac{\alpha(\,\cdot\,)\I_{\beta}[p_{\theta}]}{p_{\theta}(\,\cdot\,)}\right)\D_{\theta}f\right]
        -h_{yy}\left(x,\frac{\alpha(x)\I_{\beta}[p_{\theta}]}{p_{\theta}(x)}\right)\D_{\theta}f(x)\frac{\alpha(x)\I_{\beta}[p_{\theta}]}{p_{\theta}(x)},
    \end{align*}
    where
    \[
        \D_{\theta}f(x)=\frac{\alpha(x)}{p_{\theta}(x)}(\I_{\beta}[fp_{\theta}]-f(x)\I_{\beta}[p_{\theta}]).  
    \]
    Since
    \begin{align*}
        \D_{\theta}\psi_{\theta,j}(x)
        &=\D_{\theta}\tilde{\psi}_{\theta,j}(x)\\
        &=\frac{\alpha(x)}{p_{\theta}(x)}\left(\I_{\beta}[\tilde{\psi}_{\theta,j}p_{\theta}]-\tilde{\psi}_{\theta,j}(x)\I_{\beta}[p_{\theta}]\right)\\
        &=-\frac{C_{\theta,j}\beta(x)-\partial_{\theta_j}\log p_{\theta}(x)}{h_{yy}(x,\alpha(x)\I_{\beta}[p_{\theta}]/p_{\theta}(x))}\frac{p_{\theta}(x)}{\alpha(x)\I_{\beta}[p_{\theta}]},
    \end{align*}
    we have
    \begin{align*}
        \A_{\theta}\psi_{\theta,j}(x)
        &=-\frac{\beta(x)}{\I_{\beta}[p_{\theta}]}\I_{\alpha}\left[\frac{p_{\theta}}{\alpha}(C_{\theta,j}\beta-\partial_{\theta_j}\log p_{\theta})\right]+C_{\theta,j}\beta(x)-\partial_{\theta_j}\log p_{\theta}(x)\\
        &=-\frac{\beta(x)}{\I_{\beta}[p_{\theta}]}(C_{\theta,j}\I_{\beta}[p_{\theta}]-\E_{\theta}[\partial_{\theta_j}\log p_{\theta}(X)])+C_{\theta,j}\beta(x)-\partial_{\theta_j}\log p_{\theta}(x)\\
        &=-\partial_{\theta_j}\log p_{\theta}(x),
    \end{align*}
    which completes the proof.
\end{proof}

\section{Proofs in Section~\ref{sec:examples}}
\label{sec:ProofExample}

In this section, we provide proofs omitted from Section~\ref{sec:examples}.

\subsection{Gradient operator and score matching}
\label{sec:ProofExampleScoreMatching}

We first show that Assumption~\ref{assumption:DKernel} is satisfied.
\begin{proposition}
    \label{prp:DKernelGradient}
    Let $\L\coloneqq\nabla_x$.
    Then, Assumption~\ref{assumption:DKernel} is satisfied.
\end{proposition}

\begin{proof}
    Apply Proposition~\ref{prp:DKernelWeightedGradient} with $M(x)=I$.
\end{proof}

\begin{proof}[Proof of Proposition~\ref{prp:SteinFormContractionScoreMatching}]
    Apply Proposition~\ref{prp:SteinFormContractionWeightedGradient} with $M(x)=I$.
\end{proof}

\begin{proof}[Proof of Proposition~\ref{prp:AsymptoticEfficiencyExampleScoreMatching}]
    Let $\phi(y)\coloneqq(-\log p)^\ast(y)$.
    The gradient and Hessian of $\phi$ are denoted by $\phi_{y}(y)$ and $\phi_{yy}(y)$.
    By the standard duality relation for Legendre transforms \citep[e.g., Theorem 26.5 of][]{Rockafellar}, 
    \[
        \phi_y(-\nabla_x \log p(x))=x,\quad \forall x\in\R^p.
    \]
    Differentiating both sides yields $\phi_{yy}(-\nabla_x \log p(x))=(-\nabla_x^2 \log p(x))^{-1}$.
    
    Since $h(x,y)=-\phi(-y)$, we obtain
    \begin{align*}
    \A_{\theta}\partial_{\theta_j}\log p_{\theta}(x)
        &=-\div_x(h_{yy}(\,\cdot\,,\nabla_x\log p_{\theta}(\,\cdot\,))\nabla_x\partial_{\theta_j}\log p_{\theta})(x)\\
        &\phantom{=}-\ang{h_{yy}(x,\nabla_x\log p_{\theta}(x))\nabla_x\partial_{\theta_j}\log p_{\theta}(x),\nabla_x\log p_{\theta}(x)}\\
        &=-\div_x(-\phi_{yy}(-\nabla_x\log p_{\theta}(\,\cdot\,))\nabla_x\partial_{\theta_j}\log p_{\theta})(x)\\
        &\phantom{=}-\ang{-\phi_{yy}(-\nabla_x\log p_{\theta}(x))\nabla_x\partial_{\theta_j}\log p_{\theta}(x),\nabla_x\log p_{\theta}(x)}\\
        &=\div_x((\nabla_x^2\log p_{\theta}(\,\cdot\,))^{-1}\nabla_x\partial_{x_j}\log p_{\theta})(x)\\
        &\phantom{=}+\ang{(\nabla_x^2\log p_{\theta}(x))^{-1}\nabla_x\partial_{x_j}\log p_{\theta}(x),\nabla_x\log p_{\theta}(x)}\\
        &=\partial_{x_j}\log p_{\theta}(x)\\
        &=-\partial_{\theta_j}\log p_{\theta}(x),
    \end{align*}
    which completes the proof.
\end{proof}

\subsection{Flip operator and ratio matching}
\label{sec:ProofExampleRatioMatching}

We first show that Assumption~\ref{assumption:DKernel} is satisfied.
\begin{proposition}
    \label{prp:DKernelFlip}
    Let $\L$ be defined in \eqref{eq:FlipOperator}.
    Then, Assumption~\ref{assumption:DKernel} is satisfied.
\end{proposition}

\begin{proof}
    Apply Proposition~\ref{prp:DKernelLocalScoringRuleDiscrete} with $N(x)\coloneqq\{x^{(j)}\mid j=1,\dots,p\}$.
\end{proof}

\begin{proof}[Proof of Proposition~\ref{prp:SteinFormContractionRatioMatching}]
    Apply Proposition~\ref{prp:SteinFormContractionLocalScoringRuleDiscrete} with $N(x)\coloneqq\{x^{(j)}\mid j=1,\dots,p\}$.
\end{proof}

\begin{proof}[Proof of Proposition~\ref{prp:AsymptoticEfficiencyExampleRatioMatching}]
    Fix $\theta\in\Theta$.
    For the above choice of $h$, the corresponding Stein operator $\A_{\theta}$ takes the form
    \[
    \A_{\theta}f(x)=\sum_{j=1}^{p}\frac{p_{\theta}(x^{(j)})}{p_{\theta}(x)+p_{\theta}(x^{(j)})}(f(x^{(j)})-f(x)).
    \]
    Consider the following spaces of polynomials
    \begin{align*}
        V&\coloneqq\operatorname{span}\{ f_S \mid S\subset\{1,\dots,p\}, \# S\leq2 \},\\
        V_0&\coloneqq\operatorname{span}\{\partial_{\theta_j}\log p_{\theta}\}=\{ f \in V \mid \E_{\theta}[f(X)]=0 \},
    \end{align*}
    where $f_S(x)\coloneqq \prod_{j\in S}x_j$ and $f_{\emptyset}(x)\coloneqq1$.
    We first show that
    \begin{equation}
        \label{eq:AsymptoticEfficiencyExampleRatioMatching}
        \{\A_{\theta}f\mid f\in V\}\subset V.
    \end{equation}
    
    Define 
    \[
        H_{\theta,j}(x)\coloneqq \eta_j + \sum_{k=1}^{j-1}\eta_{kj}x_{k} + \sum_{k=j+1}^{p}\eta_{jk}x_k.
    \]
    Then
    \[
        \frac{p_{\theta}(x^{(j)})}{p_{\theta}(x)+p_{\theta}(x^{(j)})}
        =\frac{1}{1+\exp(2x_jH_{\theta,j}(x))}
        =\frac{1}{2}(1-x_j\tanh(H_{\theta,j}(x))).
    \]
    By the Fourier expansion theorem \citep[Theorem 1.1]{ODonnell2014}, $\tanh(H_{\theta,j}(x))$ is a polynomial of degree at most $p-1$.
    For $f_S\in V$, we have
    \[
        \A_{\theta}f_S(x)=-\sum_{j\in S}(f_{S}(x)-f_{S\setminus\{j\}}(x)\tanh(H_{\theta,j}(x))),
    \]
    which is a polynomial of degree at most $\max\{2,p-1\}$.
    Hence, if $p\leq3$, then $\A_{\theta}f_S\in V$, and therefore \eqref{eq:AsymptoticEfficiencyExampleRatioMatching} holds.

    Combining \eqref{eq:AsymptoticEfficiencyExampleRatioMatching} with Stein's identity yields $\{\A_{\theta}f \mid f\in V_0\}\subset V_0$.
    Hence, there exists $\Lambda(\theta)\in\R^{d\times d}$ such that 
    \[
        \A_{\theta}\partial_{\theta_j}\log p_{\theta}=\sum_{k=1}^{d}\Lambda(\theta)_{jk}\partial_{\theta_k}\log p_{\theta}.
    \]
    Since $h_{yy}(x,y)$ is negative definite, Propositions~\ref{proposition:SuffCondForAsm} and \ref{prp:DKernelFlip} imply that Assumption~\ref{assumption:SteinFormKernel} is satisfied.
    Thus, by Lemma~\ref{lem:SteinOperatorKernel}, $\operatorname{Ker}\A_{\theta}|_{V_0}=\{0\}$.
    Therefore, $\Lambda(\theta)$ is nonsingular, and the Stein score functions are given by $\psi_{\theta,j}=-\sum_{k=1}^{d}(\Lambda(\theta)^{-1})_{jk}\partial_{\theta_{k}}\log p_{\theta}$.
    The conclusion then follows from Theorem~\ref{thm:AsymptoticEfficiency}. 
\end{proof}

\subsection{Marginalization operator and Besag's pseudolikelihood}
\label{sec:ProofExamplePseudoLikelihood}

We first show that Assumption~\ref{assumption:DKernel} is satisfied.
\begin{proposition}
    \label{prp:DKernelMarginalOperator}
    Let $\L$ be defined in \eqref{eq:MarginalizationOperator}.
    Then, Assumption~\ref{assumption:DKernel} is satisfied.
\end{proposition}

\begin{proof}
    Fix an arbitrary $\theta\in\Theta$.
    Observe that
    \[
        \D_{\theta}f(x)_j=\frac{\int_{\X_j}p_{\theta}(x)\d x_j}{p_{\theta}(x)}(\E_{\theta}[f(X)\mid X_{-j}=x_{-j}]-f(x)).
    \]
    Thus, $\D_{\theta}f=0$ if and only if 
    \[
        f(x)=\E_{\theta}[f(X)\mid X_{-j}=x_{-j}]
    \]
    for all $x\in\X$ and $j=1,\dots,p$.
    Since this means that for each $j=1,\dots,p$, $f(x)$ does not depend on $x_j$, this is equivalent to $f$ being constant, which completes the proof.
\end{proof}

\begin{proof}[Proof of Proposition~\ref{prp:SteinFormContractionPseudoLikelihood}]
    Fix $\theta^\star\in\Theta$. We show that for $f_{\theta}\coloneqq\Psi(p_{\theta^\star}(\,\cdot\,)/p_{\theta}(\,\cdot\,))\in\F_{\theta}$,
    $\E_{\theta^\star}[\A_{\theta}f_{\theta}(X)]=0$ implies $\theta=\theta^\star$.
    Observe the relation
    \[
    \E_{\theta^\star}[\A_{\theta}f_{\theta}(X)]
    =-\sum_{j=1}^{p}\E_{\theta}\left[\operatorname{Cov}_{\theta}\left(\Psi\left(\frac{p_{\theta^\star}(X)}{p_{\theta}(X)}\right),\frac{p_{\theta^\star}(X)}{p_{\theta}(X)}\,\middle|\, X_{-j}\right)\right]\leq0,
    \]
    where the inequality follows from the strict monotonicity of $\Psi$.
    Equality holds if and only if, for each $j=1,\dots,p$, $p_{\theta^\star}(x)/p_{\theta}(x)$ does not depend on $x_{j}$.
    This is equivalent to $p_{\theta^\star}/p_{\theta}$ being constant, that is, $\theta=\theta^\star$, which completes the proof.
\end{proof}

\subsection{Noise-weighted integral operator and noise contrastive estimation}
\label{sec:ProofExampleNCE}

We first show that Assumption~\ref{assumption:DKernel} is satisfied.
\begin{proposition}
    \label{prp:DKernelNCE}
    Let $\L f(x)\coloneqq n(x)\int_\X f(x)\d x$.
    Then, Assumption~\ref{assumption:DKernel} is satisfied.
\end{proposition}
\begin{proof}
    Apply Proposition~\ref{prp:DKernelWeightedIntegralOperator} with $\alpha(x)=n(x),\beta(x)=1$.
\end{proof}

\begin{proof}[Proof of Proposition~\ref{prp:SteinFormContractionNCE}]
    Apply Proposition~\ref{prp:SteinFormContractionWeightedIntegral} with $\alpha(x)=n(x),\beta(x)=1$.
\end{proof}

\begin{proof}[Proof of Proposition~\ref{prp:SteinScoreNCE}]
    Apply Proposition~\ref{prp:SteinScoreWeightedIntegral} with $\alpha(x)=n(x),\beta(x)=1$ and $h(x,y)=(1+\nu y)\log(1+\nu y)-\nu y\log(\nu y)$.
\end{proof}

\subsection{Integral operator and MLE}
\label{sec:ProofExampleMLE}

We first show that Assumption~\ref{assumption:DKernel} is satisfied.
\begin{proposition}
    \label{prp:DKernelIntegralOperator}
    Let $\L f\coloneqq\int_\X f(x)\d x$.
    Then, Assumption~\ref{assumption:DKernel} is satisfied.
\end{proposition}
\begin{proof}
    Apply Proposition~\ref{prp:DKernelWeightedIntegralOperator} with $\alpha=\beta=1$.
\end{proof}

\begin{proof}[Proof of Proposition~\ref{prp:SteinFormContractionMLE}]
    Apply Proposition~\ref{prp:SteinFormContractionWeightedIntegral} with $\alpha=\beta=1$.
\end{proof}

\end{document}